%% file: main.tex
\documentclass{article}

\usepackage[preprint]{corl_2026} 

\usepackage{microtype}
\usepackage{graphicx}
\usepackage{subcaption}
\usepackage{booktabs}
\usepackage{tabularx}
\usepackage{natbib}
\usepackage{float}
\usepackage{colortbl}
\usepackage{subcaption} 
\usepackage{array}
\usepackage[table]{xcolor}
\usepackage{pgfplots}
\usepackage{wrapfig}
\usepackage{algorithm}
\usepackage{algpseudocode}
\usepackage{amsmath}
\usepackage{tabularx}
\usepackage[T1]{fontenc}
\usepackage[accsupp]{axessibility}  
\usepackage{hyperref}
\usepackage{orcidlink}
\usepackage{xspace}
\usepackage{subcaption}
\usepackage{caption}
\usepackage{adjustbox}
\usepackage{bm}

\newcommand{\modelname}{\textbf{\texttt{WIZARD}}\xspace}

\title{Robotic Policy Adaptation\\ via Weight-Space Meta-Learning}

\author{
  Christian Bianchi\footnote[2]{Correspondence
to <christian.bianch@italailabs.com>}\\
  ItalAI\\
  \And
  Siamak Yousefi\\
  ItalAI\\
  \And
  Alessio Sampieri\\
  ItalAI\\
  \And
  Andrea Roberti\\
  University of Verona\\
  \And
  Luca Rigazio\\
  ItalAI\\
  \And
  Fabio Galasso\\
  ItalAI and Sapeinza University of Rome\\
  \And
  Luca Franco\\
  ItalAI\\
}

\begin{document}
\maketitle


    
    
\begin{abstract}
Vision-Language-Action (VLA) models are emerging as a promising paradigm for robotic manipulation, enabling general-purpose policies trained from large corpora of demonstrations and action labels. However, adapting these models to new tasks still typically requires task-specific demonstrations, action annotations, and additional fine-tuning, making deployment costly and difficult to scale. 

We propose \modelname, a weight-space meta-learning framework that sidesteps task-specific fine-tuning by generating task-specific LoRA parameters for a frozen VLA policy. Given only a language instruction and a short demonstration video, \modelname predicts the corresponding adaptation weights in a single forward pass, without target-task action labels or test-time optimization. During meta-training, \modelname learns to map task evidence directly to expert LoRA updates, capturing relationships between tasks in weight space. 

Experiments on LIBERO show that \modelname improves performance by up to $\sim 2\times$ on unseen dataset collections and up to $\sim 14\times$ on unseen tasks. On a Franka Emika Panda, \modelname consistently improves over a real-domain adapted baseline, showing that generated adapters provide task-level specialization beyond simulation.

\let\thefootnote\relax\footnotetext{$\dagger$  Correspondence
to <christian.bianchi@italailabs.com>}
    
\end{abstract}


\input{sections/1_introduction}
\input{sections/2_related_works}
\input{sections/3_method}
\input{sections/4_experiments}
\input{sections/5_discussion}

\input{sections/6_conclusions}

\bibliography{main}

\newpage
\input{sections/appendix}

\end{document}

%% file: sections/1_introduction.tex


\input{figures/teaser}
\section{Introduction}
Robotic manipulation is increasingly addressed through large Vision-Language-Action (VLA) models, which reframe control as large-scale data-driven pretraining followed by task-specific adaptation~\cite{brohan2023rt1,driess2023palmeembodiedmultimodallanguage}. Despite their broad pretraining, these models often suffer a severe performance drop when deployed on unseen tasks or distribution shifts. The standard solution is to specialize the base model through task-specific fine-tuning, often using lightweight adapters such as Low-Rank Adaptation (LoRA)~\cite{hu2022lora}. While effective, this paradigm poses two key scalability issues: every new task requires task-specific demonstrations with action labels, and adaptation still requires additional fine-tuning to train a separate \textit{expert} policy~\cite{zhou2025liberopro}.

This limitation is particularly evident in modern VLA models~\cite{kim24openvla, Black20240AV, intelligence2025pi05visionlanguageactionmodelopenworld}, where the need for adaptation is drastic: the pretrained state-of-the-art $\pi_{0.5}$ achieves a 0\% success rate on unseen dataset suites such as LIBERO-Spatial. While fine-tuning on related datasets (LIBERO-Object, -Goal, -10) improves performance only marginally to 19\%, direct test-time fine-tuning on LIBERO-Spatial reaches 94\%. Although effective, such pipelines still require learning a dedicated \textit{expert} for each target distribution, raising a central question: can we adapt VLA policies to new tasks without task-specific optimization at test time?


Existing alternatives address only part of this scalability problem. Task-specific expert policies require dedicated data, optimization, and model storage for every new task. LoRA-based fine-tuning reduces the storage cost by learning lightweight adapters, but it still requires action-labeled demonstrations and additional optimization for each target task. Meta-learning could avoid repeated fine-tuning by directly generating task-specific policies, yet generating full VLA parameters is impractical for modern robotic foundation models. Recent parameter-generation methods remain limited in this respect: they rely on action-space supervision~\cite{liang2025makeanagentgeneralizablepolicynetwork,hegde2025warpdworldmodelassisted}, require privileged task information such as goal images~\cite{zhou2025hypergoalnetgoalconditionedmanipulationpolicy}, or synthesize full policy parameters, which does not scale to large VLA backbones.

We introduce \modelname\ (\textbf{W}eight-space \textbf{I}nference for \textbf{Z}ero-shot \textbf{A}daptation from \textbf{R}obotic \textbf{D}emonstration), a weight-space meta-learning framework that generates task-specific adapters on demand from a language instruction and a short demonstration video. 
Our formulation treats robotic adaptation as a parameter inference problem, where a meta-network predicts the appropriate policy update from task evidence. This differs from classic test-time adaptation, which typically relies on optimizing policies in the action space through supervision or gradient-based fine-tuning that leverages action labels.
To this end, as illustrated in Fig.~\ref{fig:teaser}, we introduce a task-oriented meta-dataset of pairs $(z, \Delta W)$, where $z$ denotes task evidence composed of language instructions and visual observations, and $\Delta W$ is the LoRA update of an expert policy trained for that task. A meta-network is then trained to map task evidence directly to expert adapter weights, learning relationships across tasks in the weight space of policy parameters. This novel formulation yields a scalable adaptation mechanism for modern state-of-the-art VLA models. 

At inference time, given only a prompt and a short video from an unseen task $z^{new}$, \modelname generates the corresponding LoRA weights $\Delta W^{new}$ in a single forward pass, without action annotations, privileged goal information, or gradient-based optimization. The generated adapter is injected into the frozen VLA backbone (i.e., $W + \Delta W^{new}$) to produce a task-specialized policy.
We evaluate the proposed framework on the LIBERO suite \cite{liu2023liberobenchmarkingknowledgetransfer}, including Goal, Object, Spatial, and 10 datasets. Compared with baselines, our method improves success rates by up to $\sim 2\times$ on average on unseen datasets and up to $\sim 14\times$ on unseen tasks, substantially narrowing the gap between pretrained VLAs and task-specific experts without test-time fine-tuning.
Our contributions are twofold:

\begin{itemize}
\item \textbf{Weight-space meta-learning for VLA adaptation.} 
    We introduce \modelname, the first weight-space meta-learning framework that generates task-specific LoRA for frozen VLA policies.

    \item \textbf{Fine-tuning-free task adaptation from task evidence.}
    Given only a language prompt and a short demonstration video, \modelname generates a task-specific adapter in a single forward pass, bypassing task-specific fine-tuning without requiring action labels or test-time optimization.

\end{itemize}

%% file: figures/teaser.tex
\begin{figure*}[h]
    \centering
    \includegraphics[width=1\textwidth]{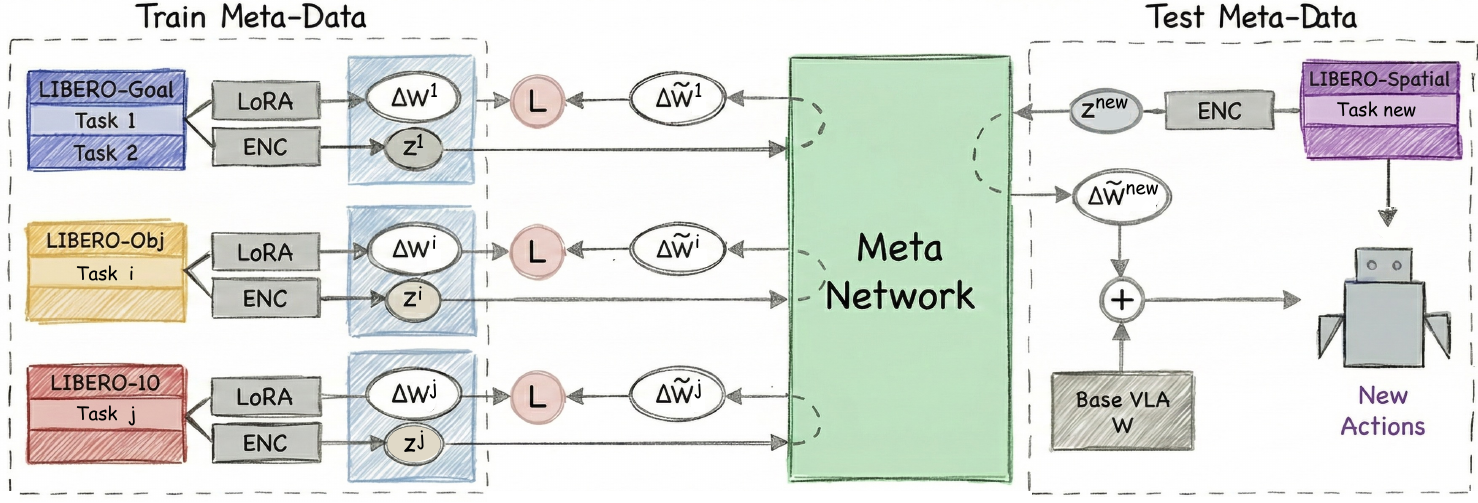}
    \caption{\textbf{WIZARD: Weight-space Inference for Zero-shot Adaptation from Robotic Demonstration.}
    \textbf{(Left) Meta-Training.} A repository of task \textit{experts} is built from LIBERO-Goal, -Object, and -10 datasets. For each task, a LoRA adapter ($\Delta W^i$) is trained while a multimodal encoder extracts a task embedding $z^i$ from the instruction and visual demonstration. The meta-network learns to map embeddings to LoRA parameters by reconstructing the expert weights ($z^i \rightarrow \Delta \tilde{W^i}$).
    \textbf{(Right) Zero-Shot Inference.} Given a new task (e.g., from LIBERO-Spatial), \modelname directly predicts the adaptation weights $\Delta \tilde{W}^{\text{new}}$ from the task embedding $z^{\text{new}}$, enabling zero-shot policy adaptation without gradient-based fine-tuning, even for unseen tasks.}
    \label{fig:teaser}
\end{figure*}

%% file: sections/2_related_works.tex
\section{Related works}
\label{sec:rel_works}

\textbf{Robotic policy adaptation and generalization.}
Generalization remains a central challenge in robotics, even with large pretrained Vision-Language-Action (VLA) models~\cite{reed2022GATO, brohan2023rt1, rt22023arxiv, driess2023palmeembodiedmultimodallanguage, kim24openvla, Black20240AV, intelligence2025pi05visionlanguageactionmodelopenworld}. While scale and diversity at training time, for example through multi-task learning such as MT-Opt~\cite{kalashnikov2021mtoptcontinuousmultitaskrobotic}, can improve robustness, policies remain constrained by the task and environment distributions in the training data, leading to failures on genuinely unseen tasks and environments. Parameter-efficient fine-tuning (PEFT) methods like LoRA~\cite{hu2022lora} enable effective specialization of these backbones~\cite{Bousmalis2023RoboCat}, by contrast \modelname does not require action-labeled task data or additional optimization at deployment.

To avoid test-time fine-tuning, training-free or low-overhead strategies such as \textit{in-context prompting} and \textit{few-shot conditioning} adapt behavior by conditioning a single policy on a small inference-time context, without updating model parameters. VIMA~\cite{jiang2023vima} exemplifies this paradigm using language and visual demonstrations, closely related to one-shot and meta-imitation approaches where tasks are inferred from minimal examples rather than fine-tuning~\cite{zhang2024oneshot}. Task2Vec~\cite{Achille2019Task2Vec} embeds visual classification tasks to estimate task similarity and transferability. TecNets~\cite{james2018tecnets} extends this idea to robotics by inferring task embeddings from demonstrations and using them to condition action generation.
More broadly, contextual and meta-RL-style methods infer latent task embeddings from limited experience to enable few-shot adaptation without per-task retraining~\cite{li2025metal, Li2025MetaVLAUM}.
A complementary line of work focuses on \emph{self-supervised adaptation during deployment}, performing online updates with unsupervised objectives to improve robustness under distribution shift. Rapid Motor Adaptation (RMA)~\cite{kumar2021rma} and PAD~\cite{guo2024prediction} follow this paradigm, but introduce test-time overhead and require carefully designed adaptation signals.

Meta-learning offers a principled alternative by leveraging relationships across tasks and datasets for fast adaptation, yet remains underexplored in robotics, particularly for large VLA policies and settings without action labels or online optimization~\cite{beck2022hypernet}. We directly address this gap by introducing a meta-learning formulation tailored to large VLA policies that enables fast adaptation without relying on action labels or requiring online optimization.

\textbf{Weight-space inference and parameter generation.}
Parameter generation aims to synthesize model weights directly rather than obtaining them solely through task-specific optimization. Early ideas appeared in fast weights~\cite{fastweight2016} and neuroevolutionary generators such as HyperNEAT~\cite{Verbancsics2013GenerativeNF}. In deep learning, HyperNetworks~\cite{ha2016hypernetworks} introduced networks that generate the weights of a target model, enabling fast model instantiation and architecture search (e.g., SMASH~\cite{Brock2017SMASHOM}) as well as parameter prediction for unseen architectures~\cite{knyazev2021parameter}. A complementary line of work treats model weights as data, learning latent spaces or generative distributions over parameters~\cite{Hyperrepresentations2022,Peebles2022}. More recent works scale this idea using diffusion-based generators that progressively denoise parameter vectors~\cite{soro2024diffusionbasedneuralnetworkweights,wang2024neural}, including conditional generators that produce efficient adaptation modules such as LoRA~\cite{jin2024conditional}. Recurrent diffusion-style generators further extend this paradigm to large models~\cite{wang2025rpg}.
A recent approach is Drag-and-Drop LLMs~\cite{liang2025draganddropllmszeroshotprompttoweights}, which learns a hyper-network that generates LoRA parameters from dataset-level embeddings using textual prompts as conditioning. However, VLA policies inherently operate over multimodal inputs combining language and visual observations. Properly accounting for this multimodal structure is therefore critical for stable and effective parameter generation in robotic settings.

In robotics, parameter generation has been explored to instantiate task- or goal-conditioned controllers. 
Hyper-GoalNet~\cite{zhou2025hypergoalnetgoalconditionedmanipulationpolicy} generates full policy parameters conditioned on privileged goal images, whereas our method predicts compact LoRA updates without requiring privileged inputs. 
More recent approaches leverage diffusion-based learning, such as Latent Weight Diffusion~\cite{hegde2025warpdworldmodelassisted} and Make-An-Agent~\cite{liang2025makeanagentgeneralizablepolicynetwork}, operate at the episode level and supervise adaptation in action space.
These approaches either generate full policy parameters or rely on privileged inputs and trajectory supervision. In contrast, we perform task-level parameter generation directly in weight space, predicting compact LoRA adapters for frozen VLA policies from language and video observations, without requiring state–action annotations or test-time optimization.

%% file: sections/3_method.tex
\section{Method}
\label{sec:method}

This section presents \modelname, our meta-learning framework for robotic VLA policies. 
We first describe the formulation of robotic adaptation as weight generation in Section~\ref{sec:method_part1}, including the construction of the meta-training dataset, the meta-network training procedure, and the zero-shot inference process.
Section~\ref{sec:method_model} introduces the key design principles required for stable training in robotics, including multimodal weight structuring, scale-aware parameter generation, and alignment-oriented supervision in weight-space.

\subsection{Meta-learning LoRA for VLA policies}
\label{sec:method_part1}

Inspired by prior works~\cite{wang2025rpg, charakorn2025texttolora, liang2025draganddropllmszeroshotprompttoweights}, we formulate robotic test-time adaptation as a \emph{weight-space generation} problem. 
Given a pretrained Vision-Language-Action model $G_W$ with frozen backbone parameters $W$, our goal is to generate task-specific parameter updates directly from task evidence, without performing gradient-based optimization at test-time.

\noindent \textbf{Task setup.}
We consider a set of manipulation tasks used for meta-training:
\begin{equation}
\mathcal{T}^{\text{train}} = \{ \tau^{(k)} \}_{k=1}^{K},
\end{equation}
where each task consists of a collection of episodes
\begin{equation}
\tau^{(k)} = \{ d_j^{(k)} \}_{j=1}^{|\tau^{(k)}|}, 
\quad
d_j^{(k)} = \big(p_j^{(k)}, v_j^{(k)}, s_j^{(k)}, a_j^{(k)} \big),
\end{equation}
with $p$ denoting the language prompt, $v$ the visual observations, $s$ the proprioceptive state, and $a$ the action sequence.
For each task $\tau^{(k)}$, we obtain an expert policy by adapting the frozen backbone $G_W$ using Low-Rank Adaptation (LoRA), producing a task-specific parameter update $\Delta W^{(k)}$.

\noindent \textbf{Meta-dataset construction.}
To construct meta-training inputs, we encode task evidence using the frozen VLA encoder itself $G^{enc}$. 
Specifically, for each task we sample a subset of episodes 
$S \subset \tau^{(k)}$ 
and compute episode embeddings
\begin{equation}
z_j^{(k)} = G^{\mathrm{enc}}\big(p_j^{(k)}, v_j^{(k)}\big).
\end{equation}
A task representation is then obtained by aggregating the encoded episodes
\begin{equation}
z^{(k)} = \frac{1}{|S|} \sum_{j \in S} z_j^{(k)}.
\end{equation}
The resulting meta-training dataset consists of task embeddings paired with the relative task expert LoRA parameters
\begin{equation}
\mathcal{D}_{\mathrm{meta}} =
\Big\{ (z^{(k)}, \Delta W^{(k)}) \;\big|\; k=1,\dots,K \Big\}.
\end{equation}
Importantly, the meta-network receives only task evidence $(p,v)$ as input, while state-action trajectories $(s,a)$ are used solely to train the expert policies.

\noindent \textbf{Meta-training.}
We train a meta-network
\begin{equation}
f: z^{(k)} \mapsto \Delta \tilde{W}^{(k)},
\end{equation}
which predicts LoRA parameter updates directly from the task representation. 
The network is optimized using a reconstruction loss in weight-space
\begin{equation}
\mathcal{L}\big(\Delta \tilde{W}^{(k)}, \Delta W^{(k)}\big).
\end{equation}

\noindent \textbf{Meta-inference.}
At evaluation time, we consider a set of unseen tasks $\mathcal{T}^{\text{test}}$, disjoint from those used for the meta-training dataset
$(\mathcal{T}^{\text{test}} \cap \mathcal{T}^{\text{train}} = \varnothing)$. 
Given new task evidence consisting of a prompt and a short visual demonstration, sampled from the test distribution $(p^{\text{new}}, v^{\text{new}}) \sim  \mathcal{T}^{\text{test}}$, we compute the task embedding
\begin{equation}
z^{\text{new}} = G^{\mathrm{enc}}(p^{\text{new}}, v^{\text{new}}),
\end{equation}
and generate the corresponding adapter parameters with our meta-network
\begin{equation}
\Delta \tilde{W}^{\text{new}} = f(z^{\text{new}}).
\end{equation}
The predicted update is applied to the frozen backbone to obtain the adapted policy $G_{W + \Delta \tilde{W}^{\text{new}}}$, enabling zero-shot execution of the new task with a single forward pass. 
Crucially, this process requires \textit{no state-action annotations} and \textit{no gradient-based optimization} at test-time.

\subsection{Design principles for meta-network training in robotics}
\label{sec:method_model}
\modelname introduces three key novelties for effective training of the meta-network in the robotic domain: 
(i) we extend weight-space meta-learning to heterogeneous VLA architectures, explicitly modeling the structural separation between perception, reasoning, and control modules; 
(ii) we introduce scale-aware parameter generation tailored to continuous robotic control, jointly predicting normalized LoRA updates and their per-layer statistics to ensure stable convergence; and 
(iii) we incorporate alignment-oriented supervision in weight space, enforcing directional and magnitude consistency between generated and expert updates beyond standard reconstruction losses.

\noindent \textbf{Multimodal weight structure for VLA architectures.}
Robotic VLA policies are architecturally heterogeneous, combining perception (vision encoder), reasoning (language transformer), and actuation (control head). As a consequence, LoRA updates span structurally distinct modules, and small perturbations in parameter magnitude lead to large behavioral deviations in continuous control. Our method explicitly accounts for this structure.
Specifically, to enable structured supervision in weight-space, each LoRA update $\Delta W^{(k)}$ is converted into a tensorized representation. Let $L$ denote the number of LoRA-adapted layers. Each layer $l$ contains low-rank matrices of rank $r$ and width $H_l$. We pad each layer to a common width $H$ and stack them into a structured tensor
\begin{equation}
\Delta W^{(k)} \in \mathbb{R}^{L \times 3 \times r \times H},
\end{equation}
where the second dimension indexes modality-specific components (vision, language, action). This representation preserves architectural boundaries while allowing joint modeling across modalities.

\noindent \textbf{Scale-aware parameter generation.}
Robotic policies are highly sensitive to parameter scale. To improve stability, we compute the representation with per-layer statistics
\begin{equation}
S^{(k)} = \{(\mu_l^{(k)}, \sigma_l^{(k)})\}_{l=1}^{L},
\end{equation}
where $\mu_l^{(k)}$ and $\sigma_l^{(k)}$ denote the mean and standard deviation of LoRA parameters in layer $l$.
The meta-network jointly predicts the weight updates and their scale
\begin{equation}
(\Delta \tilde{W}^{(k)}, \tilde{S}^{(k)}) = f(z^{(k)}),
\end{equation}
and reconstructed parameters are rescaled before being injected into the frozen VLA backbone $G_W$
\begin{equation}
\Delta \tilde{W}^{(k)} \leftarrow \tilde{S}^{(k)} \frac{\Delta \tilde{W}^{(k)}}{\|\Delta \tilde{W}^{(k)}\|}.
\end{equation}

Explicit scale modeling is necessary for stable training: without per-layer normalization, the meta-network training diverges due to the heterogeneous magnitude of multi-modal VLA parameters.

\noindent \textbf{Alignment-oriented supervision in weight space.}
To further promote robust weight prediction, we introduce directional supervision in weight space. In particular, we augment the standard reconstruction objective with a cosine-based loss that enforces directional consistency between generated and expert parameter updates. This term encourages the meta-network to preserve the functional direction of the expert policies in weight space, beyond simply matching parameter values.\\
The overall training objective combines three components:
\begin{equation}
\mathcal{L}
=
\mathcal{L}_{\text{MSE}}
+
\lambda_{\text{scale}} \mathcal{L}_{\text{scale}}
+
\lambda_{\text{cos}} \mathcal{L}_{\text{cos}},
\label{eq:loss_total}
\end{equation}
where
\begin{equation}
\mathcal{L}_{\text{MSE}} = \|\Delta W^{(k)} - \Delta \tilde{W}^{(k)}\|^2,
\label{eq:loss_mse}
\end{equation}
\begin{equation}
\mathcal{L}_{\text{scale}} = \|S^{(k)} - \tilde{S}^{(k)}\|^2,
\label{eq:loss_scale}
\end{equation}
\begin{equation}
\mathcal{L}_{\text{cos}} = 1 - \mathrm{CosSim}\Bigg(
\frac{\Delta W^{(k)}}{\|\Delta W^{(k)}\|},
\frac{\Delta \tilde{W}^{(k)}}{\|\Delta \tilde{W}^{(k)}\|}
\Bigg).
\label{eq:loss_cos}
\end{equation}
The cosine term promotes directional alignment in weight space, while the scale term enforces accurate magnitude reconstruction. Together with the MSE reconstruction loss, these objectives encourage both structural and functional consistency between generated and expert policies.

%% file: sections/4_experiments.tex
\section{Experiments}
\label{sec:exp}
We show that robotic policy parameters lie on a structured, learnable weight manifold, and that traversing this space enables zero-shot generalization to novel spatial configurations without test-time fine-tuning. Our experiments evaluate weight-space inference against supervised fine-tuning, retrieval-based adaptation, and task-specific expert policies under a strict held-out distribution shift. Section~\ref{sec:exp_setup} describes the experimental protocol and benchmarks, Section~\ref{sec:exp_quant} reports quantitative comparisons across all LIBERO datasets, Section~\ref{sec:rw_exp} reports real-world results, and Section~\ref{sec:disc} analyzes data efficiency and warm-start adaptation.

\subsection{Experimental setup}
\label{sec:exp_setup}

\noindent \textbf{LIBERO benchmark.}
To evaluate test-time generalization under controlled distributional shifts, we conduct experiments on the \textbf{LIBERO} benchmark suite~\cite{liu2023liberobenchmarkingknowledgetransfer}, which consists of four complementary datasets: \textit{LIBERO-Spatial} (layout variations), \textit{LIBERO-Object} (object-centric manipulation), \textit{LIBERO-Goal} (goal-conditioned behaviors), and \textit{LIBERO-10} (long-horizon multi-stage tasks). 
These datasets share similar environments while inducing distinct task distributions.

\noindent \textbf{Meta-network training and inference.}
We adopt a \textit{held-out} distribution protocol: the meta-network is trained on three LIBERO datasets (e.g., Object, Goal, 10) and evaluated on the remaining unseen dataset (e.g., Spatial). At inference time, the meta-network generates LoRA parameters for tasks drawn from a distribution never observed during meta-training, without any gradient updates.

\noindent \textbf{Generated policy evaluation.}
All experiments follow the official implementation of $\pi_{0.5}$~\cite{intelligence2025pi05visionlanguageactionmodelopenworld} using the \texttt{openpi} infrastructure and the LIBERO benchmark, which leverages the MuJoCo physics engine. Generated policies are evaluated on 50 predefined initial states per task to ensure reproducible evaluation under distribution shifts. We report the \textit{Success Rate}, averaged over the 50 evaluations.


\noindent\textbf{Baselines.}
We compare \modelname\ with four reference settings reported in Table~\ref{tab:main_results}: task-specific $\pi_{0.5}$ LoRA experts as an in-distribution upper bound, a nearest-neighbor (NN) baseline that retrieves and applies the closest training-task adapter, and two MT-VLA baselines fine-tuned on the three meta-training datasets using OpenVLA-OFT and $\pi_{0.5}$ backbones. Unlike these baselines, which rely on supervised training or retrieval from existing adapters, \modelname\ generates a new adapter from unlabeled task evidence without test-time optimization.

\input{tables/main_results_short}

\subsection{Quantitative results}
\label{sec:exp_quant}
We evaluate zero-shot performance across all LIBERO datasets under a strict held-out distribution shift protocol. Table~\ref{tab:main_results} reports the success rates for each held-out dataset.

\noindent \textbf{Zero-shot task adaptation.}
The pretrained $\pi_{0.5}$ VLA without task-specific adaptation consistently achieves $0\%$ success across all tasks and is omitted from the table for brevity. The zero-shot Nearest-Neighbor baseline remains consistently low across datasets, with non-zero success on only a few tasks.
Our weight-generation approach substantially improves over the MT-VLA baselines while remaining strictly zero-shot. 
In LIBERO-Spatial, \modelname reaches 0.40 average success, more than twice the success rate of MT-VLA ($0.19$).
Similar gains appear in LIBERO-Goal and LIBERO-Object, where WIZARD improves the average success rate from $0.14$ to $0.22$ and from $0.01$ to $0.03$, respectively.


\paragraph{Task-level adaptation behavior.}
Across distribution shifts, \modelname shows that generated LoRA adapters recover task-specific behaviors rather than collapsing to generic policies. In LIBERO-Spatial, our method recovers the required spatial priors, achieving strong performance on several tasks, such as Task 1 ($0.90$) and Task 3 ($0.82$). Under the severe visual shifts of LIBERO-Object, the MT-VLA baselines frequently fail, whereas our generated adapters recover measurable performance on tasks such as Task 5 and Task 10. In LIBERO-Goal, generated weights effectively encode task-specific affordances, reaching $0.86$ on Tasks 5 and 9. Importantly, \modelname also recovers non-trivial performance on difficult tasks where the MT-VLA baselines remain near-zero, including LIBERO-Spatial (Tasks 2, 5, 6, 7, 9), LIBERO-Object (Tasks 2, 5, 10), and LIBERO-Goal (Tasks 3, 6, 7, 8, 10), highlighting the ability of weight-space generation to recover task-specific structure beyond standard zero-shot baselines.

Finally, LIBERO-10 introduces sequential subtask composition (A/B). While full-task zero-shot completion remains at $0.00$, subtask-level metrics indicate improved isolated skill execution compared to MT-VLA, suggesting that the main limitation lies in compositional sequencing rather than individual skill acquisition.

\begin{figure}[b]
    \centering
    \includegraphics[width=1\textwidth]{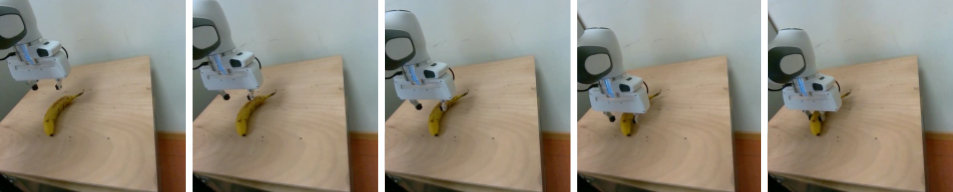}
    \caption{\textbf{Real-world qualitative rollout.} 
Example execution of \modelname\ on the real Franka Emika Panda for the \textit{pick the banana} task, progressing from approach to grasp.}
    \label{fig:real_qualitative}
    \vspace{-0.5cm}
\end{figure}

\subsection{Real-world experiment}
\label{sec:rw_exp}

We further evaluate \modelname on a real 7-DoF Franka Emika Panda equipped with a parallel gripper. Visual observations are acquired from three Intel RealSense cameras: one D415 mounted in an eye-in-hand configuration and two external D405 cameras, as shown in Fig.~\ref{fig:real_world_setup}. All cameras are calibrated using the method of~\citet{zhang2000flexible}. The VLA policy outputs commands at $15$~Hz, which are upsampled through a velocity interpolator to continuous $1$~kHz setpoints tracked by a low-level joint impedance controller.

We consider five real-world manipulation tasks: \textit{pick the banana}, \textit{pick the apple}, \textit{pick the marker}, \textit{pick the cup}, and \textit{move the apple close to the cup}. Directly deploying the pretrained $\pi_{0.5}$ backbone leads to near-zero performance due to the sim-to-real gap, with the robot often drifting away from the target workspace. Therefore, both methods start from a $\pi_{0.5}$ checkpoint pretrained on DROID~\citep{droid_2024} and lightly adapted to the target real setup using $30$ episodes. This protocol does not test strict zero-shot sim-to-real transfer, but evaluates whether \modelname can improve task-level adaptation once the base VLA is grounded in the real robot domain.

Fig.~\ref{fig:real_qualitative} shows a real-world banana rollout, from approach to grasp. 
Table~\ref{tab:real_world_results} reports real-world success rates computed over 30 evaluation trials per task. 
\modelname\ improves over the real-domain adapted $\pi_{0.5}$ baseline on all tasks, increasing the average success rate from $0.22$ to $0.41$. 
The largest gains appear on \textit{banana} ($0.27 \rightarrow 0.53$), \textit{apple} ($0.13 \rightarrow 0.33$), and \textit{cup} ($0.30 \rightarrow 0.63$), while smaller but consistent improvements are observed on \textit{marker} ($0.10 \rightarrow 0.17$) and \textit{apple $\rightarrow$ cup} ($0.07 \rightarrow 0.17$). 
Since both methods share the same real-domain initialization, these results suggest that the gains come from weight-space task adaptation rather than from the initial sim-to-real calibration alone.

\input{tables/real_world_results}




%% file: tables/main_results_short.tex
\begin{table*}[t]
\centering

\scriptsize
\setlength{\tabcolsep}{2.2pt}
\renewcommand{\arraystretch}{0.92}

\resizebox{\textwidth}{!}{%
\begin{tabular}{l|cccccccccc|c||cccccccccc|c}
\toprule
\multicolumn{1}{c}{} 
& \multicolumn{11}{c||}{\textbf{LIBERO-Spatial}}
& \multicolumn{11}{c}{\textbf{LIBERO-Object}} \\
\cmidrule(lr){2-12}\cmidrule(lr){13-23}
 & T1 & T2 & T3 & T4 & T5 & T6 & T7 & T8 & T9 & T10 & \textbf{Avg.}
 & T1 & T2 & T3 & T4 & T5 & T6 & T7 & T8 & T9 & T10 & \textbf{Avg.} \\
\midrule
$\pi_{0.5}$ Experts
& 1.00 & 0.98 & 1.00 & 0.94 & 0.92 & 0.98 & 0.96 & 0.98 & 0.96 & 0.96 & 0.97
& 0.96 & 0.98 & 0.98 & 0.96 & 0.96 & 0.98 & 0.96 & 0.96 & 1.00 & 0.98 & 0.97 \\
\midrule
Nearest-Neighbor (NN) & 0.00 & 0.02 & 0.08 & 0.12 & 0.00 & 0.00 & 0.04 & 0.00 & 0.00 & 0.00 & 0.02
& 0.00 & 0.00 & 0.00 & 0.00 & 0.02 & 0.00 & 0.00 & 0.00 & 0.00 & 0.00 & 0.00 \\
MT-VLA (OpenVLA-OFT)
& 0.18 & 0.00 & 0.36 & 0.58 & 0.00 & 0.00 & 0.00 & 0.04 & 0.00 & 0.00 & 0.09
& 0.02 & 0.00 & 0.00 & 0.00 & 0.00 & 0.00 & 0.00 & 0.00 & 0.00 & 0.02 & 0.00 \\
MT-VLA ($\pi_{0.5}$)
& 0.22 & 0.00 & 0.56 & \textbf{0.86} & 0.00 & 0.02 & 0.00 & 0.18 & 0.02 & 0.00 & 0.19
& \textbf{0.10} & 0.02 & 0.00 & 0.00 & 0.02 & 0.00 & 0.00 & 0.00 & 0.00 & 0.00 & 0.01 \\
\midrule
\rowcolor{gray!15}
\modelname
& \textbf{0.90} & \textbf{0.12} & \textbf{0.82} & 0.84 & \textbf{0.08} & \textbf{0.28} & \textbf{0.10} & 0.76 & \textbf{0.08} & 0.00 & \textbf{0.40}
& 0.08 & \textbf{0.06} & 0.00 & 0.00 & \textbf{0.06} & 0.00 & 0.00 & 0.00 & 0.00 & \textbf{0.08} & \textbf{0.03} \\

\bottomrule
\end{tabular}%
}

\vspace{0.45cm}

\scriptsize
\setlength{\tabcolsep}{2.0pt}
\renewcommand{\arraystretch}{0.92}

\resizebox{\textwidth}{!}{%
\begin{tabular}{l|cccccccccc|c||cccccccccc|c}
\toprule
\multicolumn{1}{c}{} 
& \multicolumn{11}{c||}{\textbf{LIBERO-Goal}}
& \multicolumn{11}{c}{\textbf{LIBERO-10 (A/B)}} \\
\cmidrule(lr){2-12}\cmidrule(lr){13-23}
 & T1 & T2 & T3 & T4 & T5 & T6 & T7 & T8 & T9 & T10 & \textbf{Avg.}
 & T1 & T2 & T3 & T4 & T5 & T6 & T7 & T8 & T9 & T10 & \textbf{Avg.} \\
\midrule
$\pi_{0.5}$ Experts
& 0.92 & 0.98 & 1.00 & 0.86 & 0.94 & 0.92 & 0.94 & 1.00 & 0.94 & 0.76 & 0.93
& 0.90 & 0.98 & 0.92 & 0.92 & 0.94 & 0.90 & 0.90 & 0.92 & 0.78 & 0.74 & 0.89 \\
\midrule
Nearest-Neighbor (NN) & 0.00 & 0.04 & 0.00 & 0.00 & 0.06 & 0.00 & 0.00 & 0.00 & 0.08 & 0.00 & 0.02
& 0.00/0.00 & 0.00/0.02 & 0.04/0.00 & 0.00/0.00 & 0.00/0.00 & 0.00/- & 0.02/0.00 & 0.00/0.02 & 0.00/0.00 & 0.00/0.00 & 0.01/0.00 \\
MT-VLA (OpenVLA-OFT) & 0.00 & 0.12 & 0.00 & 0.00 & 0.18 & 0.00 & 0.02 & 0.00 & 0.20 & 0.00 & 0.05
& 0.00/0.00 & 0.00/0.00 & 0.08/0.00 & 0.00/0.00 & 0.00/0.02 & 0.00/- & 0.04/0.00 & 0.00/0.06 & 0.00/0.00 & 0.00/0.00 & 0.01/0.01 \\
MT-VLA ($\pi_{0.5}$)
& 0.00 & \textbf{0.40} & 0.02 & 0.00 & 0.46 & 0.00 & 0.04 & 0.00 & 0.46 & 0.00 & 0.14
& 0.02/0.02 & 0.02/0.02 & 0.24/0.00 & 0.02/0.00 & 0.00/\textbf{0.06} & 0.00/- & 0.12/0.00 & 0.00/0.24 & 0.00/0.00 & 0.00/0.00 & 0.03/0.03 \\
\midrule
\rowcolor{gray!15}
\modelname
& 0.00 & 0.08 & \textbf{0.10} & 0.00 & \textbf{0.86} & \textbf{0.06} & \textbf{0.12} & \textbf{0.06} & \textbf{0.86} & \textbf{0.08} & \textbf{0.22}
& \textbf{0.08}/\textbf{0.06} & \textbf{0.08}/\textbf{0.08} & \textbf{0.34}/0.00 & \textbf{0.08}/0.00 & \textbf{0.06}/\textbf{0.06} & 0.00/- & \textbf{0.20}/\textbf{0.06} & 0.00/\textbf{0.30} & \textbf{0.06}/\textbf{0.04} & 0.00/\textbf{0.08} & \textbf{0.09}/\textbf{0.07} \\

\bottomrule
\end{tabular}%
}

\caption{\textbf{Zero-shot task-conditioned weight generation across LIBERO benchmarks.} Each block evaluates held-out distribution shift, with the NN and the MT-VLA baselines, and generates per-task LoRA adapters with \modelname. 
Expert rows denote task-specific fine-tuning upper bounds.}
\label{tab:main_results}
\end{table*}

%% file: tables/real_world_results.tex
\begin{figure}[!h]
\centering

\adjustbox{valign=c}{%
\begin{minipage}{0.65\linewidth}
\centering
\setlength{\tabcolsep}{3.5pt}
\renewcommand{\arraystretch}{1.05}
\resizebox{\linewidth}{!}{
\begin{tabular}{l|ccccc|c}
\toprule
\textbf{Method} 
& \textbf{Banana} 
& \textbf{Apple} 
& \textbf{Marker} 
& \textbf{Cup} 
& \textbf{Apple $\rightarrow$ Cup} 
& \textbf{Avg.} \\
\midrule
$\pi_{0.5}$ + real-domain adaptation 
& 0.27 & 0.13 & 0.10 & 0.30 & 0.07 & 0.17 \\
\rowcolor{gray!15}
\modelname
& 0.53 & 0.33 & 0.17 & 0.63 & 0.17 & 0.33 \\
\bottomrule
\end{tabular}
}
\vspace{0.35em}
\captionof{table}{\textbf{Real-world evaluation on Franka Emika.} Success rates on five real-world manipulation tasks.}
\label{tab:real_world_results}
\end{minipage}
}%
\hfill%
\adjustbox{valign=c}{%
\begin{minipage}{0.30\linewidth}
\centering
\includegraphics[width=\linewidth]{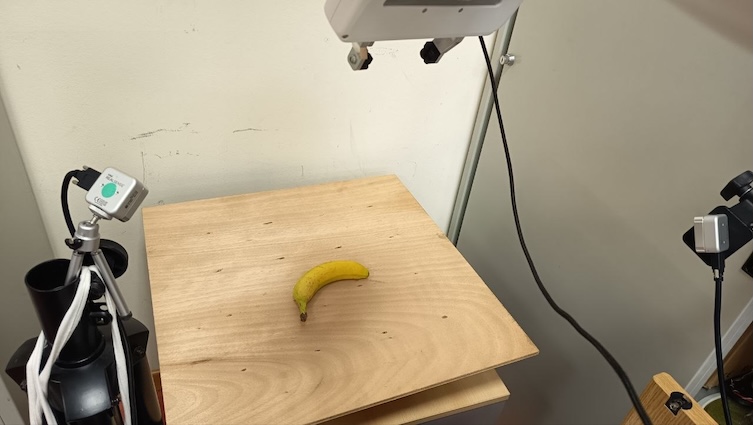}
\vspace{-0.3cm}
\captionof{figure}{\textbf{Real-world setup.} 
}
\label{fig:real_world_setup}
\end{minipage}
}
\vspace{-.3cm}
\end{figure}

%% file: sections/5_discussion.tex
\subsection{Analysis and discussion}
\label{sec:disc}
\vspace{-.2cm}
Beyond zero-shot execution, we test whether generated adapters reduce supervision and accelerate fine-tuning when adaptation is needed.

\noindent \textbf{Data efficiency.}
We compare \modelname\ with MT-VLA policies fine-tuned with increasing numbers of task-specific demonstrations on LIBERO-Spatial Task 1. As shown in Fig.~\ref{fig:efficiency}, \modelname\ reaches 90\% success without task-specific gradient updates, while MT-VLA starts from 22\% and needs about 25 demonstrations to match it. Although fine-tuning with all 50 demonstrations approaches expert performance, it requires additional labeled data and optimization. This highlights that \modelname\ can synthesize effective task-specific adapters directly from task evidence.

\noindent \textbf{Warm-start adaptation.}
We also test whether generated adapters help when fine-tuning is still allowed. On LIBERO-Spatial Task 10, where both \modelname\ and MT-VLA fail zero-shot, initializing from the \modelname-generated weights speeds up convergence, reaching the 96\% expert success rate in 70 steps instead of 90 (Fig.~\ref{fig:efficiency_warmstart}). This suggests that even unsuccessful zero-shot adapters can place the policy in a better region for rapid adaptation.

\input{figures/efficiency_comparison_new}

%% file: figures/efficiency_comparison_new.tex
\begin{figure}[!h]
\centering

\definecolor{eccvcyan}{RGB}{0,150,150}
\definecolor{eccvorange}{RGB}{230,90,40}
\definecolor{expertgray}{RGB}{100,100,100}

\begin{subfigure}[t]{0.48\linewidth}
\centering
\begin{tikzpicture}
\begin{axis}[
    width=\linewidth,
    height=5.2cm,
    xmin=0, xmax=50,
    ymin=0, ymax=100,
    xtick={0,1,2,5,10,25,50},
    ytick={0,20,40,60,80,90,100},
    yticklabels={0,20,40,60,80,\textcolor{eccvcyan}{90},100},
    xlabel={Episodes},
    xlabel style={font=\small, yshift=.2cm},
    ylabel={Success Rate (\%)},
    ylabel style={font=\small, yshift=-.5cm},
    legend pos=south east,
    ymajorgrids=true,
    grid style={dashed, gray!30},
    tick label style={font=\scriptsize},
    label style={font=\small},
    legend style={
        font=\scriptsize,
        draw=gray!30,
        fill=white,
        inner sep=.5pt,
        row sep=.5pt
    }
]

\addplot[
    color=eccvcyan,
    mark=*,
    mark size=2pt,
    only marks
] coordinates {(0,90)};
\addlegendentry{\modelname}

\addplot[
    color=eccvcyan,
    dashed,
    line width=1.2pt,
    domain=0:50,
    forget plot
] {90};

\addplot[
    color=eccvorange,
    mark=*,
    mark size=1.5pt,
    line width=1.2pt,
] coordinates {
(0,18)(1,22)(2,52)(5,72)(10,80)(25,90)(50,94)
};
\addlegendentry{MT-VLA}

\draw [black!60, densely dotted, thick] (axis cs:25,0) -- (axis cs:25,90);
\node[circle, fill=eccvcyan, inner sep=2pt, draw=white, line width=1pt] at (axis cs:25,90) {};

\end{axis}
\end{tikzpicture}
\caption{Data efficiency comparison.}
\label{fig:efficiency}
\end{subfigure}
\hfill
\begin{subfigure}[t]{0.48\linewidth}
\centering
\begin{tikzpicture}
\begin{axis}[
    width=\linewidth,
    height=5.2cm,
    xmin=0, xmax=100,
    ymin=0, ymax=100,
    xtick={0,10,20,30,40,50,60,70,80,90,100},
    ytick={0,20,40,60,80,96,100},
    yticklabels={0,20,40,60,80,\textbf{96},100},
    xlabel={Steps},
    xlabel style={font=\small, yshift=.2cm},
    ylabel={Success Rate (\%)},
    ylabel style={font=\small, yshift=-.5cm},
    legend pos=south east,
    ymajorgrids=true,
    grid style={dashed, gray!30},
    tick label style={font=\scriptsize},
    label style={font=\small},
    legend style={
        font=\scriptsize,
        draw=gray!30,
        fill=white,
        inner sep=.5pt,
        row sep=.5pt
    }
]

\addplot[
    color=expertgray,
    dashed,
    line width=1.2pt,
    domain=0:100,
] {96};
\addlegendentry{Expert}

\addplot[
    color=eccvorange,
    mark=square*,
    mark size=1.5pt,
    line width=1.2pt,
    smooth
] coordinates {
    (0,0)(10,30)(20,50)(30,64)(40,74)(50,82)(60,87)(70,91)(80,94)(90,96)(100,96)
};
\addlegendentry{MT-VLA}

\addplot[
    color=eccvcyan,
    mark=*,
    mark size=1.5pt,
    line width=1.2pt,
    smooth
] coordinates {
    (0,0)(10,50)(20,70)(30,82)(40,89)(50,93)(60,95)(70,96)(80,96)(90,96)(100,96)
};
\addlegendentry{\modelname}

\draw [black!60, densely dotted, thick] (axis cs:70,0) -- (axis cs:70,96);
\node[circle, fill=eccvcyan, inner sep=2pt, draw=white, line width=1pt] at (axis cs:70,96) {};

\draw [black!60, densely dotted, thick] (axis cs:90,0) -- (axis cs:90,96);
\node[circle, fill=eccvorange, inner sep=2pt, draw=white, line width=1pt] at (axis cs:90,96) {};

\end{axis}
\end{tikzpicture}
\caption{Warm-start adaptation.}
\label{fig:efficiency_warmstart}
\end{subfigure}

\caption{\textbf{Adaptation efficiency.}
\textbf{(a)} \modelname\ matches 25-demo MT-VLA.
\textbf{(b)} Generated weights warm-start fine-tuning and reach expert performance faster.}
\label{fig:analysis_efficiency}
\vspace{-.3cm}
\end{figure}
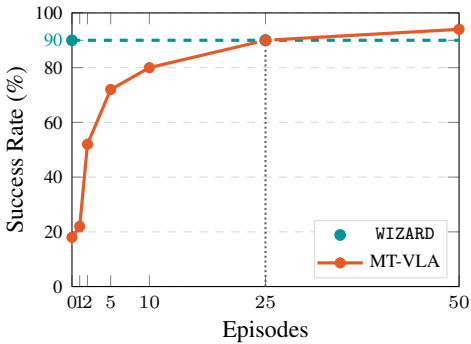
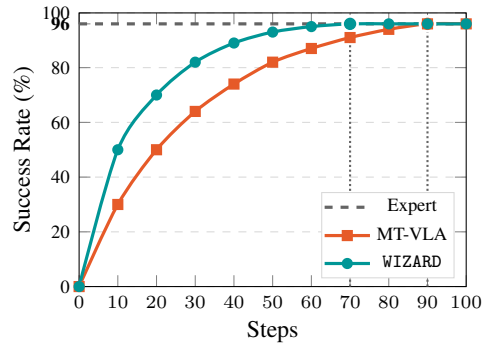

%% file: sections/6_conclusions.tex
\section{Conclusions}
\label{sec:conclusions}
\vspace{-.3cm}
We introduced \modelname, a weight-space meta-learning framework that generates LoRA adapters for frozen VLA policies from language and video evidence. \modelname\ enables zero-shot adaptation without action labels or test-time optimization, showing strong generalization, data efficiency, and effective warm-starts on LIBERO.

\noindent\textbf{Limitations.}
\modelname\ depends on the quality of the task evidence: ambiguous or incomplete demonstrations may produce unreliable adapters. It also targets single-task expert generation, limiting performance on long-horizon compositional tasks such as LIBERO-10. Future work will explore richer task representations and expert composition.


%% file: sections/appendix.tex
\appendix
\section*{Appendix}
This appendix provides supplementary material supporting the main paper. 
Appendix~\ref{sec:app_B} reports additional analyses and ablations, including embedding structure, conditioning granularity, support size, modality ablations, loss components, intra-dataset generalization, and scaling costs. 
Appendix~\ref{sec:app_C} presents extended qualitative rollouts across LIBERO. 
Appendix~\ref{sec:app_D} lists the semantic prompts and task definitions used for evaluation. 
Appendix~\ref{sec:app_E} provides implementation details, including expert LoRA training, parameter tokenization, meta-network architecture, optimization, and the topological sorting rulebook.

We also provide supplementary videos in the accompanying \texttt{index.html} file.



\section{Additional analysis}
\label{sec:app_B}

We provide additional analyses supporting the main results. 
Specifically, we study: 
(i) the topological structure of task embeddings and generated weights, 
(ii) the effect of dataset-level versus task-level conditioning, 
(iii) the impact of support size when constructing task embeddings, 
(iv) the contribution of visual and textual modalities, 
(v) the role of each training loss component, 
(vi) cross-task generalization within datasets, and 
(vii) speed and memory scaling.

\noindent \textbf{Topological structure of task embeddings and weights.}
We first analyze the structure of the conditioning space using t-SNE visualizations of the task embeddings $z_i$ (Fig.~\ref{fig:tsne_combined}). Across all LIBERO datasets (ref. Fig.~\ref{fig:tsne_combined}, left), embeddings form well-separated clusters corresponding to both individual tasks and dataset families, indicating that tasks behave as isolated experts while still lying on a structured manifold.
Focusing on LIBERO-Spatial (ref. Fig.~\ref{fig:tsne_combined}, middle), embeddings obtained from prompt and video demonstrations remain strongly separated across tasks, showing that the multimodal encoder already captures task-level semantics before weight generation. Fig.~\ref{fig:tsne_combined} (right) visualizes the latent representation at the final layer of the meta-network, approximating the latent space of generated parameters. Although spread more widely, the relative task structure is preserved: isolated tasks (e.g., Tasks 4 and 6) remain separated, while similar tasks move even closer to each other (e.g., Tasks 1 and 5; Tasks 9 and 10). This suggests that the meta-network learns a weight representation aligned with the geometry of the input embedding space. The task-number mapping is reported in the Appendix.

\input{figures/tsne_plots}

\noindent \textbf{Dataset embedding vs. task embedding.}
We analyze the effect of embedding granularity by comparing dataset-level and task-level conditioning. Prior work~\cite{liang2025draganddropllmszeroshotprompttoweights} computes dataset embeddings by averaging episodes across all tasks in a dataset. Instead, we compute embeddings using only episodes from a single task.
Table~\ref{tab:ablation_grid} (left) shows that the dataset-level conditioning achieves 0.11 average success, below $\pi_{0.5}$ (0.19). Task-level conditioning reaches 0.40, an improvement of 29 percentage points with respect to dataset-level performance. Dataset-level embeddings are therefore too coarse and collapse task structure, whereas task-level embeddings preserve the semantics required for weight generation.

\input{tables/discussion_all}

\noindent \textbf{Impact of support size on task embeddings.}
We study how performance varies with the number of episodes used to build the task embedding. We vary the support size $|S|$ by averaging embeddings from randomly sampled episodes of the same task. Table~\ref{tab:ablation_grid} (right) shows that performance remains largely stable across support sizes. This indicates that even a single episode provides sufficiently informative task evidence to generate a useful LoRA adapter. This is encouraging, since it suggests that \modelname can operate with minimal test-time evidence.\\
At the same time, increasing the number of support episodes does not lead to a clear improvement, although some individual tasks benefit from additional samples (e.g., T6 improves from $0.08$ at $|S|=1$ to $0.38$ at $|S|=5$). We hypothesize that this limited scaling may be due to the frozen $\pi_{0.5}$ text and video encoders used to construct task embeddings, whose representations may not be sufficiently robust or general-purpose for aggregating diverse visual evidence. Future work will investigate stronger task encoders trained on broader and more diverse data, potentially with representation-learning objectives such as contrastive learning, to obtain more robust task embeddings as the support size increases.

\noindent \textbf{Visual grounding vs. semantic instruction.}
\modelname conditions only on text prompts and video frames rather than state-action datasets. To evaluate the role of each modality, we ablate them independently.
Removing video conditioning causes a complete collapse in performance (0\%), while video-only conditioning achieves 18\% success, compared to 33\% with full multimodal input. This suggests that text provides the high-level task intent (\textit{what}), whereas visual demonstrations provide the geometric and kinematic information required for execution (\textit{how}). Without visual grounding, the meta-network lacks the spatial information needed to generate effective policy parameters.

\noindent \textbf{Ablation on the training loss.}
In Table~\ref{tab:loss_abl}, we evaluate the contribution of each component of the training objective (see Eqs. 14, 15, 16, 17) by progressively removing the proposed loss terms. 
In the first row, we observe that using the standard reconstruction objective $\mathcal{L}_{\text{MSE}}$ only leads to a complete failure of the meta-network (0.00 average success), indicating that a pure parameter regression objective is insufficient for robotic VLA policies. 
Similarly, in the second row, combining $\mathcal{L}_{\text{MSE}}$ with the cosine alignment loss $\mathcal{L}_{\text{cos}}$ still results in 0.00 average performance, highlighting that directional alignment alone cannot compensate for scale inconsistencies across LoRA parameters. 
Introducing the scale-aware objective $\mathcal{L}_{\text{scale}}$  (third row) unlocks the effectiveness of the framework, increasing the average success rate to 0.27. 
This behavior is expected in the robotic setting, where LoRA parameters can vary significantly in magnitude across tasks and across layers, making explicit scale prediction necessary for stable weight generation. 
Finally, in the last row, we show the performance by adding the cosine alignment loss. The combination of the three losses further improves the average performance from 0.27 to 0.33. 
This additional term encourages directional consistency between generated and expert updates, helping organize task adapters within the weight manifold and improving overall policy performance.

\input{tables/table_loss_abl}




\noindent \textbf{Cross-task generalization within datasets.}
We study intra-dataset generalization by holding out two tasks within each LIBERO dataset and using them as test set.
We fine-tune $\pi_{0.5}$ to obtain the MT-VLA baseline and train \modelname on 32 tasks (8 per dataset), evaluating both methods on the held-out tasks. Table~\ref{tab:task_based_training} shows high success rates for both methods on LIBERO-Goal and LIBERO-Object (e.g., 100\% vs 98\% on Goal Task 5; 98\% vs 84\% on Object Task 10). On LIBERO-Spatial and LIBERO-10, MT-VLA performs slightly better (e.g., 94\% vs 90\% on Spatial Task 6; 80\% vs 66\% on LIBERO-10 Task 1), although our method remains competitive.
While gradient-based fine-tuning remains effective when the test tasks are similar to the annotated ones during supervised fine tuning phase, our method generalizes to unseen tasks without supervision and performs better in the cross-dataset setting (ref. Table~\ref{tab:main_results}).\\

\input{tables/task_based_training}

\noindent \textbf{Speed and Memory analysis.}
Tab.~\ref{tab:rank_scaling_cost} reports a rank-scaling analysis on an NVIDIA A6000 GPU with 48GB of memory. The main bottleneck is memory rather than runtime: generation remains fast across ranks, while memory increases from 1.74GB at $r=16$ to 23.5GB at $r=512$, and full-model generation reaches out-of-memory. These results show that generating full VLA policies is impractical at scale, whereas \modelname remains feasible by generating compact LoRA updates.

\input{tables/speed_memory}

\clearpage

\section{Additional qualitative results}
\label{sec:app_C}
We present additional qualitative rollouts across the LIBERO benchmark and real-world setup to further illustrate the behavior of the policies generated by \modelname. These examples highlight the ability of the generated adapters to generalize across spatial reasoning, object recognition, goal-conditioned manipulation, and atomic actions within composed tasks and real-world deployment.\\

\noindent \textbf{LIBERO-Spatial.}
In Figs.~\ref{fig:add_qualitative_spatial} and~\ref{fig:add_qualitative_spatial2}, we report additional qualitative rollouts for the tasks with non-zero success rate (all except Task 10). Task 8 is already shown in the main paper. 

These examples further highlight the ability of the generated policies to recover precise kinematic priors and adapt the end-effector trajectory to reach atypical spatial configurations. Across the reported tasks, the robot successfully retrieves the target bowl from different positions in the scene and places it on the plate, demonstrating consistent spatial reasoning and collision-aware manipulation.\\

\noindent \textbf{LIBERO-Object.}
In Fig.~\ref{fig:add_qualitative_object}, we report qualitative rollouts for the tasks with non-zero success rate, such as Tasks 1, 2, and 5 (Task 10 is shown in the main paper). 

These results emphasize the visual grounding capabilities of the generated policies: the robot correctly identifies novel objects such as the alphabet soup, cream cheese, and ketchup among distractors and executes the appropriate pick-and-place actions, confirming robust object-level generalization under appearance variation.\\

\noindent \textbf{LIBERO-Goal.}
In Fig.~\ref{fig:add_qualitative_goal}, we report qualitative results for the tasks with non-zero success rate (all except Tasks 1 and 4), while Task 10 is presented in the main paper. 

The examples illustrate the generation of task-specific affordances driven by the language instruction, including actions such as placing objects on the stove or cabinet, inserting items into containers, and activating the stove. These rollouts highlight the ability of \modelname\ to synthesize specialized manipulation primitives directly from task descriptions.\\

\noindent \textbf{LIBERO-10.}
LIBERO-10 tasks involve sequential compositions of two manipulation skills. For visualization clarity, we report in Figs.~\ref{fig:add_qualitative_10} and~\ref{fig:add_qualitative_102} the qualitative rollouts by separating each task into its two constituent stages, highlighting the execution of each sub-action independently. This allows us to inspect how the generated policy decomposes the overall objective into intermediate steps, while we will research as future work how to successfully complete the composed behavior.

\noindent \textbf{Real-world experiment.}
In Fig.~\ref{fig:add_qualitative_real}, we report additional qualitative rollouts from the real-world Franka Emika setup. The examples show three atomic manipulation tasks used in our real evaluation: \textit{pick up the apple}, \textit{pick up the banana}, and \textit{pick up the cup}. Across these rollouts, \modelname\ is able to ground the target object from the language instruction, approach it from the available camera observations, and execute the corresponding grasping behavior on physical hardware. Compared to simulation, these executions are more sensitive to perception noise, calibration errors, grasp alignment, and contact dynamics, which explains the larger gap between simulated and real-world success rates. Nevertheless, the qualitative examples show that the generated adapters can transfer to real manipulation and produce meaningful task-specific behaviors beyond synthetic benchmarks.

\clearpage

\input{figures/app_qualitative_spatial}
\input{figures/app_qualitative_spatial2}

\input{figures/app_qualitative_object}

\input{figures/app_qualitative_goal}

\input{figures/app_qualitative_10}
\input{figures/app_qualitative_102}
\input{figures/app_qualitative_real}

\clearpage

\section{Additional details on the real-world setup}
\label{sec:app_C2}

The experiments were conducted using a Franka Emika Panda manipulator, a collaborative robotic arm featuring seven revolute joints and seven degrees of freedom (7-DoF). The redundant kinematic structure provides enhanced dexterity and enables the robot to maintain favorable configurations while executing complex manipulation tasks. The Panda integrates joint torque sensing at each actuator and exposes low-level torque control interfaces, making it particularly suitable for interaction-rich applications requiring compliant behavior. Its lightweight mechanical design and backdrivable joints further contribute to safe operation in environments involving physical contact.

To ensure compliant motion throughout the manipulator, a joint-space impedance controller was employed. The control law is based on inverse-dynamics feedback linearization \cite{siciliano_textbook}
\begin{subequations}\label{eq:imped_control}
        \begin{equation}
            \bm{\tau} = \bm{B}(\bm{q}) \bm{y} + \bm{n}(\bm{q}, \dot{\bm{q}}) + \bm{\tau}_e
        \end{equation}
        followed by
        \begin{equation}
            \bm{y} = \bm{M}^{-1}\bigl( \bm{D}(\dot{\bm{q}}_d - \dot{\bm{q}}) + \bm{K}(\bm{q}_d - \bm{q}) - \bm{\tau}_e \bigr) + \ddot{\bm{q}}_d
        \end{equation}
\end{subequations}

Here, $\bm{\tau}\in\mathbb{R}^{7}$ denotes the commanded joint torques, while $\bm{q}$, $\dot{\bm{q}}$, and $\ddot{\bm{q}}$ represent the joint positions, velocities, and accelerations, respectively. The vectors $\bm{q}_d$, $\dot{\bm{q}}_d$, and $\ddot{\bm{q}}_d$ correspond to the desired joint-space trajectory. The matrix $\bm{B}(\bm{q})\in\mathbb{R}^{7\times7}$ is the manipulator inertia matrix, whereas $\bm{n}(\bm{q},\dot{\bm{q}})\in\mathbb{R}^{7}$ collects the Coriolis, centrifugal, and gravitational terms. The vector $\bm{\tau}_e\in\mathbb{R}^{7}$ represents the measured external joint torques. The diagonal positive-definite matrices $\bm{M}$, $\bm{D}$, and $\bm{K}\in\mathbb{R}^{7\times7}$ define the desired impedance mass, damping, and stiffness, respectively.

The desired end-effector trajectory is first converted into the corresponding joint-space reference through the analytical inverse kinematics formulation developed for the Franka Emika Panda manipulator \cite{HeLiu2021}. The resulting joint references are then tracked by the impedance controller, yielding compliant behavior along the entire kinematic chain while preserving accurate trajectory execution.

Visual observations are acquired using three Intel RealSense depth cameras. A RealSense D415 is mounted directly on the robot end-effector in an eye-in-hand configuration, providing a viewpoint that follows the manipulator motion and enables close-range perception of the workspace. The D415 combines a high-resolution RGB sensor with an active stereo depth sensing system, delivering synchronized color and depth information suitable for object localization, scene reconstruction, and visual servoing applications. To complement the eye-in-hand observations, two externally mounted RealSense D405 cameras are positioned around the workspace to provide additional viewpoints and reduce occlusions. Unlike the D415, the D405 is specifically designed for high-accuracy short-range depth sensing and is particularly effective for capturing fine geometric details at close distances.

All cameras are intrinsically calibrated using the method proposed by \citet{zhang2000flexible}. For each camera, multiple images of a planar checkerboard calibration target are acquired at different orientations and positions within the field of view. Corner features are automatically detected in each image and used to estimate the camera intrinsic parameters, including focal lengths, principal point coordinates, and lens distortion coefficients. The calibration process minimizes the reprojection error between the observed image points and the corresponding projected points predicted by the camera model.

\subsection{Teleoperated Data Acquisition}
Teleoperated demonstrations were collected to build a dataset of task executions under human supervision. The data acquisition was performed using the native Cartesian impedance controller provided by the Franka Emika interface, which enables compliant end-effector motion by regulating both position and orientation in task space. In this setting, the robot behaves as a mass--spring--damper system in Cartesian coordinates, allowing the operator to physically guide the end-effector while preserving safe and stable interaction with the environment.

\begin{figure}
    \centering
    \includegraphics[width=0.5\linewidth]{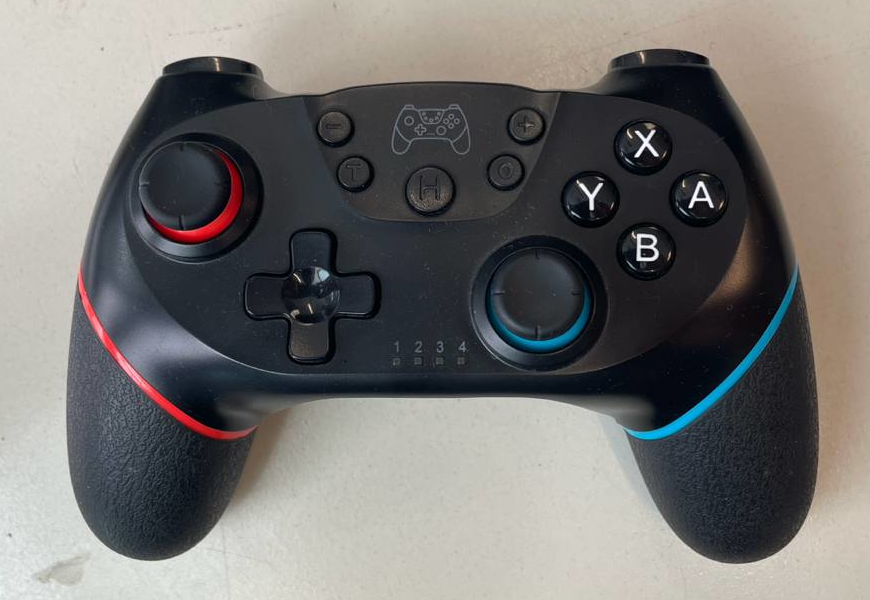}
    \caption{Handheld input device used for robot teleoperation during data acquisition.}
    \label{fig:app_controller}
\end{figure}

The controller was driven through a standard teleoperation interface, where the human operator issued motion commands using a handheld input device, shown in Figure~\ref{fig:app_controller}. The commanded inputs were mapped to Cartesian velocity and pose increments, which were then tracked by the impedance controller in real time. This setup allowed the operator to directly guide the end-effector trajectory while the low-level impedance behavior ensured smooth motion, attenuation of abrupt user inputs, and consistent contact stability during task execution.
The resulting demonstrations consist of synchronized streams of robot states, including joint positions, velocities, and applied torques, together with the corresponding Cartesian end-effector trajectories.

\subsection{ROS-based System Integration for Distributed Inference}
The system architecture relies on \texttt{ROS1} to enable communication between the workstation performing policy inference and the low-level robot controller. This setup decouples perception and decision-making from real-time robot execution, allowing the inference module to run on a dedicated computing unit while maintaining synchronized interaction with the Franka Emika Panda control stack.
Robot state and sensory observations are streamed through standard ROS topics, which are used as the primary interface between modules. In particular, joint states are received via \texttt{/joint\_states},while gripper state information is obtained from \texttt{/franka\_gripper/joint\_states}. Visual inputs are acquired from multiple RGB streams, including the wrist-mounted camera and two external cameras, published on \texttt{/camera/color/image\_raw}, \texttt{/camera\_left/color/image\_raw}, and \texttt{/camera\_right/color/image\_raw}, respectively. These observations are synchronized and packaged into a structured state representation that is passed to the inference pipeline.

Policy inference is executed on a remote client using a WebSocket-based communication interface, which receives the aggregated observation and returns a chunked sequence of predicted actions. The returned actions include both joint-space commands for the 7-DoF manipulator and a scalar signal for gripper control. The action chunk is executed in an open-loop fashion over a short horizon, after which a new inference request is issued to update the control trajectory.

The computed joint commands are published back to the robot through a dedicated ROS topic (\texttt{/vla/joint\_commands}), ensuring compatibility with the existing control stack. Gripper actions are executed through the standard \texttt{franka\_gripper} action interface, which provides separate \texttt{move} and \texttt{grasp} action servers for open and close commands. This modular design enables a clean separation between perception, policy inference, and low-level actuation, while maintaining real-time operation and consistent synchronization between sensing and control.

\clearpage

\section{Semantic prompts and task definitions}
\label{sec:app_D}

To ensure full reproducibility and to explicitly define the semantic support of our meta-training and evaluation manifolds, we detail the exact prompt instructions utilized for zero-shot conditioning. The $\pi_{0.5}$ foundation model relies on these specific string inputs to project the visual observations into the task latent space.\\

\noindent \textbf{LIBERO-Spatial}
The spatial dataset evaluates the model's ability to ground relative positional geometry. Notably, Tasks 5 and 10 introduce the token \texttt{`wooden cabinet'}, an unseen concept that triggered the total performance collapse.

\input{tables/task_table}

\clearpage

\section{Extended implementation details}
\label{sec:app_E}

In this section, we provide an exhaustive breakdown of the meta-network's ($\pi_{0.5}$) architecture, the deterministic LoRA parameter tokenization pipeline, the formulation of the multi-objective loss function, and the exact distributed training dynamics.\\

\noindent \textbf{Two-stage LoRA fine-tuning and variance generation}
To construct the diverse dataset of target weights required to train the meta-network, we employ a novel two-stage Low-Rank Adaptation (LoRA) fine-tuning pipeline on the $\pi_{0.5}$ foundation model. This pipeline is executed using PyTorch Distributed Data Parallel (DDP) across multiple GPUs, utilizing \texttt{bfloat16} mixed precision to optimize memory overhead.\\

\textit{Phase 1: Domain-specific expert training.} 
In the first phase, our objective is to anchor the base $\pi_{0.5}$ model to a specific, high-performing domain expert manifold. We freeze the base model and inject trainable LoRA adapters ($r=16, \alpha=16, \text{dropout}=0.0$) across all linear target modules, initialized via a standard Gaussian distribution.

The model is trained exclusively on a single manipulation suite to isolate specific grounding capabilities. Optimization is performed using the 8-bit AdamW optimizer with $\beta_1=0.9, \beta_2=0.999$, and gradient clipping at a maximum norm of $1.0$. We employ a cosine decay learning rate schedule, warming up over 25 steps to a peak learning rate of $5 \times 10^{-5}$, and decaying smoothly. This phase runs until the model reaches a performance plateau (approximately 10,000 steps), yielding a stable ``Expert'' parameter state.

\textit{Phase 2: High-frequency variance generation.} 
Once the expert plateau is reached, we freeze the global trajectory and pivot to variance generation. The goal of this phase is to harvest a dense, diverse set of functional LoRA weights that represent slightly different, highly capable expert configurations for the same task. This ensemble captures the local variance of the optimization landscape around the expert minimum, serving as the meta-training dataset for the meta-network. We initialize this phase by directly resuming the optimizer state and LoRA weights from the Phase 1 expert checkpoint. To prevent catastrophic forgetting of the domain prior and ensure the generated weights remain strict task experts, the optimization dynamics are drastically constrained: the peak learning rate is reduced to $1 \times 10^{-5}$ and decays aggressively to $1 \times 10^{-7}$ over a micro-horizon of just 500 steps.\\

\noindent \textbf{Data processing and LoRA tokenization}
The foundation model consists of highly heterogeneous modules, including a Vision Transformer (PaliGemma), a language expert (Gemma), and various multi-modal projectors. To construct a unified latent space for the meta-network, we introduce \texttt{Pi05LoRATokenizer2D}, which deterministically maps 1D LoRA weight vectors into standardized 2D token grids.\\

\textit{Padding and chunking:} 
All target LoRA parameters are constrained to a fixed 2D token resolution of $16 \times 512$. Because the underlying weight matrices vary drastically in size (e.g., the language model head contains up to $257,536$ parameters per projection), the tokenizer applies asymmetric chunking. Weights are partitioned along the input dimension into $N_{\text{tokens}}$ distinct chunks. If a dimension is not perfectly divisible by the token resolution, it is zero-padded up to the nearest multiple (e.g., the \texttt{mlp.fc1} layer in the vision tower is padded to $1536$ and $4608$ for the up and down projections, respectively). In total, the tokenizer maps the full parameter set into exactly $5,929$ discrete tokens per training episode.\\

\textit{Normalization and scale tracking:}
To stabilize the generative manifold, we apply instance-wise normalization to each parameter chunk before tokenization. While the explicit scale tracking parameters and their corresponding objective ($\mathcal{L}_{\text{scale}}$) are established in Section 3.2, we also implement a spatial boundary heuristic for zero-shot scale reconstruction in the event of missing scale conditioning. During detokenization, pseudo-scales are inferred by taking the expectation over the trailing $2 \times 2$ boundary elements along both the height and width of the $16 \times 512$ token grid. These reconstructed scales are then averaged and adjusted using a fixed dataset prior ($\text{mean\_scale} = 8.0$, $\text{std\_bias} = 1.6$) before un-normalizing the tokens back into the continuous parameter space.\\

\textit{Manifold stabilization via sorting:}
To prevent permutation-induced mode collapse during meta-training, we enforce a strict, absolute topological sorting of the parameter keys. Tensors are sorted by functional priority: Action Projectors $\rightarrow$ Gemma Expert $\rightarrow$ PaliGemma Language Model $\rightarrow$ Multi-Modal Projector $\rightarrow$ Vision Tower $\rightarrow$ Temporal MLPs.\\

\noindent \textbf{Meta-network architecture}
\label{appendix:architecture}
The generative backbone is a custom 3D fully convolutional network designed to process sequence-aligned Visual-Language-Action (VLA) embeddings.\\

\textit{Feature alignment:} 
The raw VLA embeddings, extracted at a dimensionality of $2048$, are first projected into a $512$-dimensional bottleneck via a linear layer. To handle variable-length kinematic demonstrations, the temporal sequence length is linearly interpolated to a fixed spatial dimension of $16$. This aligned feature map is then temporally tiled across $168$ distinct time steps, yielding an input tensor of shape $168 \times 16 \times 512$.\\

\textit{Decomposed 3D convolutional blocks:} 
The core of the decoder utilizes custom 3D convolutional layers. Rather than computing a standard 3D convolution, which is prohibitively expensive and prone to overfitting, we decompose the operation into parallel dual-pathways. Let $X$ be the input feature map. The layer computes a width-first convolution $X_1$ and a height-first convolution $X_2$:
$$X_1 = \text{Conv}_{h}(\text{Conv}_{w}(X)), \quad X_2 = \text{Conv}_{w}(\text{Conv}_{h}(X))$$
Specifically, $\text{Conv}_{w}$ applies a $1 \times 7$ spatial kernel, while $\text{Conv}_{h}$ applies a $7 \times 1$ spatial kernel. Both operations maintain the $512$-channel dimensionality without introducing an intermediate bottleneck. The parallel pathways $X_1$ and $X_2$ are then averaged, shifted by a learned bias, passed through a SiLU activation, and regularized with a Dropout probability of $p=0.15$.\\

\textit{Dimensionality progression:} 
The decoder gradually expands the temporal dimension while compressing and then expanding the spatial features to generate the final $5,929$ tokens:
\begin{enumerate}
    \item \textbf{Layer 0 (Input):} $168 \times 16 \times 512$
    \item \textbf{Layer 1 (Compression):} $128 \times 32 \times 512$
    \item \textbf{Layer 2 (Processing):} $128 \times 32 \times 512$
    \item \textbf{Layer 3 (Pre-Expansion):} $1024 \times 16 \times 512$
    \item \textbf{Layer 4 (Output Expansion):} $5929 \times 16 \times 512$
\end{enumerate}

\textit{Scale predictor:} 
Operating in parallel to the decoder, the scale predictor maps the flattened $16 \times 512$ aligned embeddings ($8192$ dimensions) through a $1024$-dimensional bottleneck. This sub-network utilizes a SiLU activation and is regularized with Dropout ($p=0.15$) before a final linear projection outputs the $4$-parameter explicit scale vectors (tracking $\mu$ and $\sigma$ for both up and down projections) for all $5,929$ tokens.\\

\noindent \textbf{Objective formulation}
\label{appendix:objective}
As formulated in Section 3.2, the model is optimized using a composite objective function ($\mathcal{L}_{\text{total}}$). However, because the tokenized grids contain variable-length padded regions, the primary reconstruction loss ($\mathcal{L}_{\text{MSE}}$) is implemented as a masked Mean Squared Error. We define $\Omega$ as the set of valid (non-padded) spatial indices within the token grids. Furthermore, to balance the learning dynamics across highly heterogeneous architectural modules, the MSE is scaled by a dataset-specific criterion weight matrix $\mathcal{W}$, derived from the element-wise standard deviation of the valid tokens across the meta-training distribution. The scale ($\mathcal{L}_{\text{scale}}$) and cosine ($\mathcal{L}_{\text{cos}}$) losses are computed exclusively over these valid indices.

\noindent \textbf{Distributed data streaming and modality slicing}
\label{appendix:streaming}
Given the extreme I/O overhead of loading thousands of parameter \texttt{.safetensors} per epoch, we employ a custom streaming dataset. Checkpoint-embedding pairs are pre-allocated to isolated worker processes to prevent file-locking bottlenecks.

Furthermore, to study the isolated effects of distinct semantic modalities, our pipeline enforces hard temporal slicing on the $2048$-dimensional sequence-aligned Visual-Language-Action (VLA) embeddings prior to feature alignment. For language-only conditioning, we extract sequence tokens $[0, 512]$; for vision-only conditioning, we isolate tokens $[512, 1280]$. When training the fully multi-modal baseline, the entire $[0, 1280]$ token sequence is utilized. Continuous data augmentation is applied dynamically at the worker level by injecting Gaussian noise ($\sigma = 0.05$) directly into these sliced embedding tensors. \\

\noindent \textbf{Optimization and training dynamics}
\label{appendix:optimization}
We train the model using Hugging Face Accelerate with mixed-precision (\texttt{bf16}) across a distributed setup. To minimize memory overhead during the generation of the massive $5,929 \times 16 \times 512$ parameter space, we employ the 8-bit AdamW optimizer.  

During training, we apply input augmentation to the VLA condition embeddings by injecting continuous Gaussian noise ($\sigma = 0.05$). The full suite of optimization hyperparameters is detailed in Table~\ref{tab:hyperparameters}.\\

\begin{table}[h]
\centering
\caption{Meta-network optimization parameters}
\label{tab:hyperparameters}
\begin{tabular}{@{}ll@{}}
\toprule
\textbf{Hyperparameter} & \textbf{Value} \\ \midrule
Optimizer & 8-bit AdamW \\
Learning Rate & $1.0 \times 10^{-4}$ \\
Weight Decay & $0.1$ \\
Learning Rate Schedule & Cosine Decay \\
Warmup Steps & $500$ \\
Total Training Steps & $5,000$ \\
Batch Size (per GPU) & $6$ \\
Gradient Accumulation Steps & $4$ \\
Effective Batch Size & $24 \times N_{GPUs}$ \\
Max Gradient Norm & $1.0$ \\
Dropout Rate & $0.15$ \\
Input Augmentation Noise & $\mathcal{N}(0, 0.05)$ \\
Mixed Precision & \texttt{bfloat16} \\ \bottomrule
\end{tabular}
\end{table}

\noindent \textbf{Topological parameter sorting and deterministic rulebook}
\label{appendix:sorting_rulebook}

A core contribution of our tokenization strategy is the enforcement of a strict topological ordering over the foundation model's parameter space.  Because the generative process relies on spatial convolutions over the tokenized weights, the sequence in which parameters are presented directly dictates the structural manifold learned by the meta-network. Random or lexicographical sorting leads to severe mode collapse, as functionally dependent layers become spatially disjointed.

To resolve this, we implement a priority-based sorting heuristic (Algorithm~\ref{alg:topological_sort}) that groups parameters by their architectural modality and functional depth.

\begin{algorithm}[h]
\caption{Topological Sorting of LoRA Parameters}
\label{alg:topological_sort}
\textbf{Input:} A LoRA parameter key string $K$ \\
\textbf{Output:} A priority tuple $(P_{primary}, P_{secondary})$ for absolute sorting
\begin{algorithmic}[1]
\State Extract semantic parts from $K$ separated by \texttt{"."}

\If{\texttt{".action\_in\_proj"} $\in K$}
    \State \Return $(0, 0)$
\ElsIf{\texttt{".action\_out\_proj"} $\in K$}
    \State \Return $(0, 1)$
\ElsIf{\texttt{".gemma\_expert.lm\_head"} $\in K$}
    \State \Return $(0, 2)$
\ElsIf{\texttt{".gemma\_expert.model.layers"} $\in K$}
    \State $L \gets$ Extract layer index from $K$
    \State \Return $(1, L + \text{ComponentOffset}(K))$
\ElsIf{\texttt{".gemma\_expert.model.norm"} $\in K$}
    \State \Return $(2, 0)$
\ElsIf{\texttt{".paligemma.lm\_head."} $\in K$}
    \State \Return $(2, 1)$
\ElsIf{\texttt{".paligemma.model.language\_model.layers"} $\in K$}
    \State $L \gets$ Extract layer index from $K$
    \State \Return $(3, L + \text{ComponentOffset}(K))$
\ElsIf{\texttt{".multi\_modal\_projector."} $\in K$}
    \State \Return $(4, 0)$
\ElsIf{\texttt{".vision\_tower."} $\in K$}
    \State $L \gets$ Extract layer index from $K$
    \State \Return $(5, L + \text{ComponentOffset}(K))$
\ElsIf{\texttt{"time\_mlp\_in"} $\in K$}
    \State \Return $(6, 0)$
\ElsIf{\texttt{"time\_mlp\_out"} $\in K$}
    \State \Return $(6, 1)$
\Else
    \State \textbf{raise} \texttt{RuntimeError}("Unexpected layer topology")
\EndIf

\end{algorithmic}
\end{algorithm}

Once the parameters are topologically sorted, they are mapped into the unified $16 \times 512$ token grids. Because the raw weight matrices possess highly irregular dimensionalities (ranging from small action projectors to massive language model heads), we apply asymmetric chunking and zero-padding.

Table~\ref{tab:tokenization_rulebook} exhaustively details the tokenization hyperparameters. The \textit{Up Chunks} and \textit{Down Chunks} denote the number of discrete $16 \times 512$ grids allocated to the $\mathbf{A}$ and $\mathbf{B}$ LoRA matrices, respectively.\\

\begin{table}[h]
\centering
\caption{Complete LoRA tokenization rulebook for $\pi_{0.5}$. The total token count across all layers dictates the $5,929$ discrete tokens generated by the meta-network.}
\label{tab:tokenization_rulebook}
\resizebox{\textwidth}{!}{%
\begin{tabular}{@{}lcccc@{}}
\toprule
\textbf{Target Component Pattern} & \textbf{Up Chunks} & \textbf{Down Chunks} & \textbf{Pad Up} & \textbf{Pad Down} \\ \midrule
\texttt{.action\_in\_proj} & 1 & 16 & - & - \\
\texttt{.action\_out\_proj} & 16 & 1 & - & - \\
\texttt{.gemma\_expert.lm\_head} & 2 & 503 & - & 257,536 \\
\texttt{.gemma\_expert...input\_layernorm.dense} & 2 & 6 & - & - \\
\texttt{.gemma\_expert...mlp.down\_proj} & 8 & 2 & - & - \\
\texttt{.gemma\_expert...mlp.gate\_proj} & 2 & 8 & - & - \\
\texttt{.gemma\_expert...mlp.up\_proj} & 2 & 8 & - & - \\
\texttt{.gemma\_expert...post\_attention\_layernorm} & 2 & 6 & - & - \\
\texttt{.gemma\_expert...self\_attn.k\_proj} & 2 & 1 & - & - \\
\texttt{.gemma\_expert...self\_attn.o\_proj} & 4 & 2 & - & - \\
\texttt{.gemma\_expert...self\_attn.q\_proj} & 2 & 4 & - & - \\
\texttt{.gemma\_expert...self\_attn.v\_proj} & 2 & 1 & - & - \\
\texttt{gemma\_expert.model.norm.dense} & 2 & 6 & - & - \\
\texttt{.paligemma...mlp.down\_proj} & 32 & 4 & - & - \\
\texttt{.paligemma...mlp.gate\_proj} & 4 & 32 & - & - \\
\texttt{.paligemma...mlp.up\_proj} & 4 & 32 & - & - \\
\texttt{.paligemma...self\_attn.k\_proj} & 4 & 1 & - & - \\
\texttt{.paligemma...self\_attn.o\_proj} & 4 & 4 & - & - \\
\texttt{.paligemma...self\_attn.q\_proj} & 4 & 4 & - & - \\
\texttt{.paligemma...self\_attn.v\_proj} & 4 & 1 & - & - \\
\texttt{.paligemma\_with\_expert.paligemma.lm\_head} & 4 & 503 & - & 257,536 \\
\texttt{.multi\_modal\_projector.} & 3 & 4 & 1,536 & - \\
\texttt{.vision\_tower...mlp.fc1} & 3 & 9 & 1,536 & 4,608 \\
\texttt{.vision\_tower...mlp.fc2} & 9 & 3 & 4,608 & 1,536 \\
\texttt{.vision\_tower...self\_attn.k\_proj} & 3 & 3 & 1,536 & 1,536 \\
\texttt{.vision\_tower...self\_attn.out\_proj} & 3 & 3 & 1,536 & 1,536 \\
\texttt{.vision\_tower...self\_attn.q\_proj} & 3 & 3 & 1,536 & 1,536 \\
\texttt{.vision\_tower...self\_attn.v\_proj} & 3 & 3 & 1,536 & 1,536 \\
\texttt{.time\_mlp} & 2 & 2 & - & - \\ \bottomrule
\end{tabular}%
}
\end{table}

%% file: figures/tsne_plots.tex
\begin{figure*}[!h]
    \centering
    
    \begin{subfigure}[t]{0.31\textwidth}
        \centering
        \includegraphics[width=\linewidth]{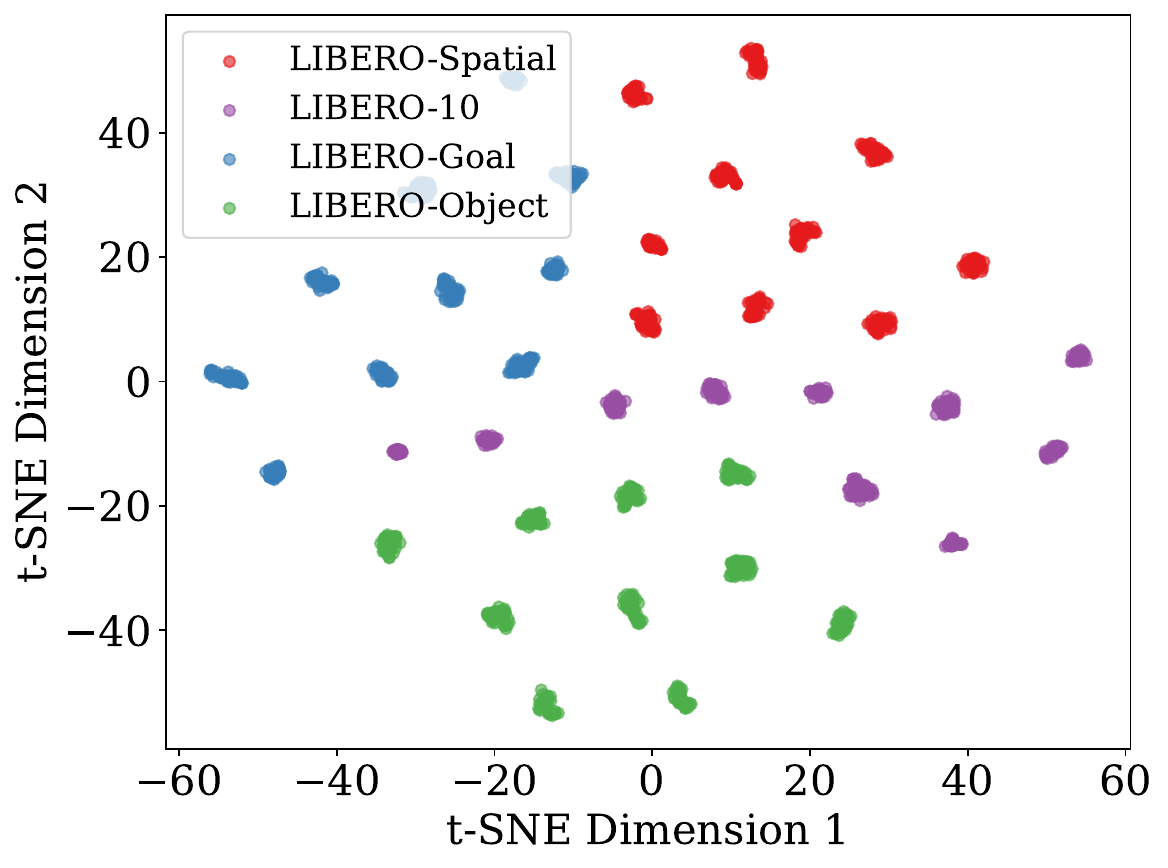}
    \end{subfigure}
    \hfill
    \begin{subfigure}[t]{0.32\textwidth}
        \centering
        \includegraphics[width=\linewidth]{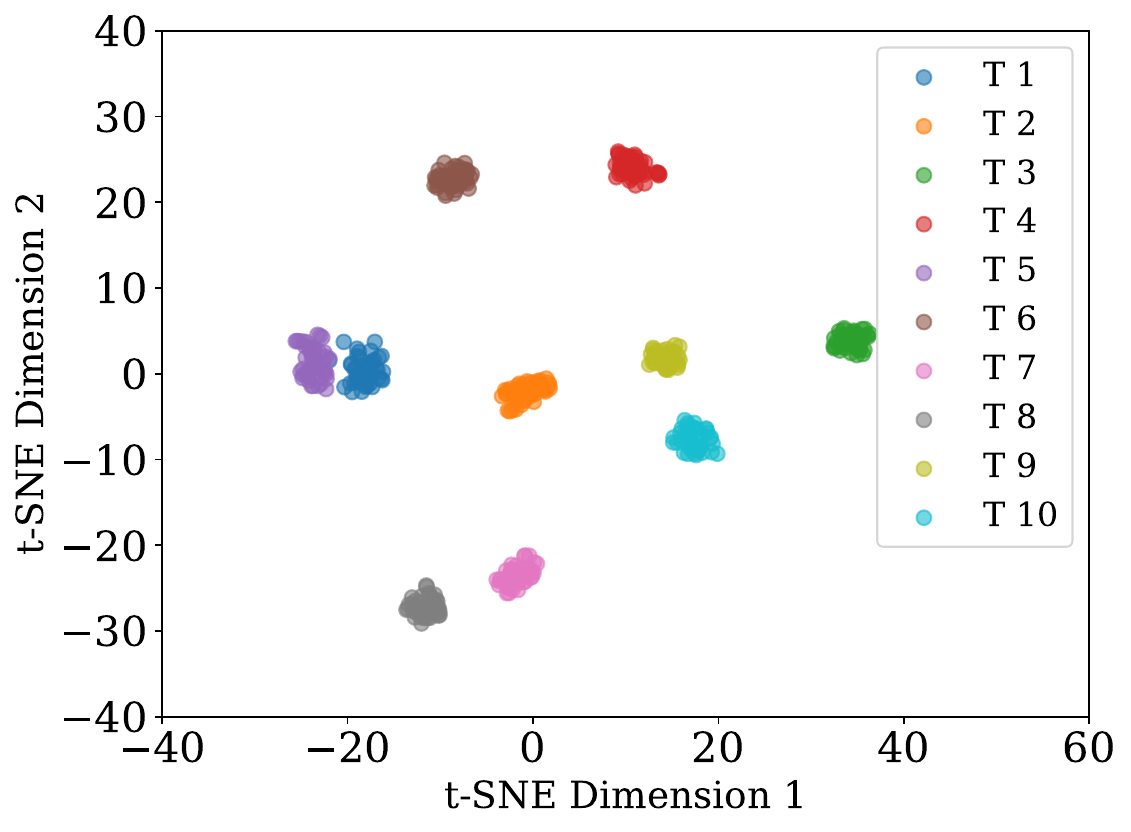}
    \end{subfigure}
    \hfill
    \begin{subfigure}[t]{0.32\textwidth}
        \centering
        \includegraphics[width=\linewidth]{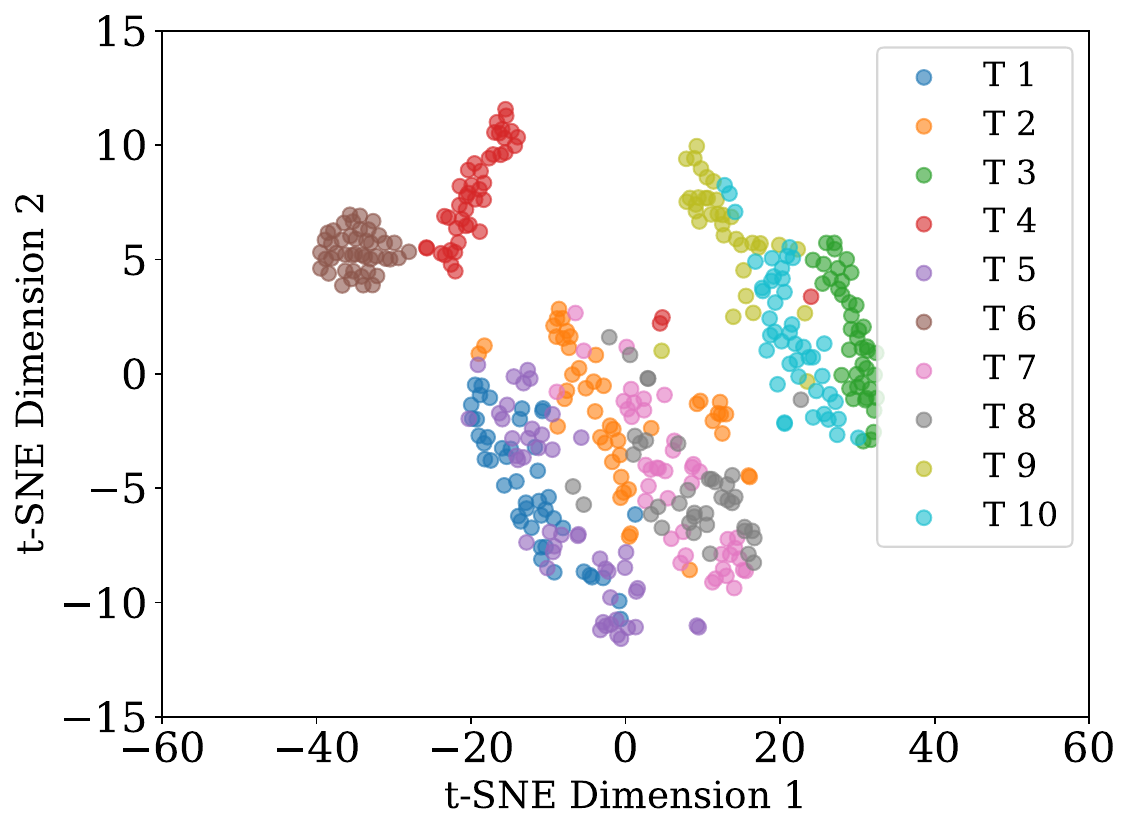}
    \end{subfigure}
    
    \caption{\textbf{Structure of the conditioning and latent spaces.}
The t-SNE visualizations of (left) task embeddings $z_i$ across all LIBERO suite, (middle) embeddings for LIBERO-Spatial tasks, and (right) the latent representation at the final layer of the meta-network. The plots show that task embeddings form distinct task-level clusters and are transformed into a structured representation in weight space.}
    
    \label{fig:tsne_combined}
\end{figure*}

%% file: tables/discussion_all.tex
\begin{table*}[!h]
\centering

\scriptsize
\setlength{\tabcolsep}{2.8pt}
\renewcommand{\arraystretch}{1.0}

\begin{subtable}[t]{0.54\textwidth}
\centering
\resizebox{\linewidth}{!}{
\begin{tabular}{l|cccccccccc|c}
\toprule
 & T1 & T2 & T3 & T4 & T5 & T6 & T7 & T8 & T9 & T10 & Avg \\
\midrule
MT-VLA ($\pi_{0.5}$)
& 0.22 & 0.00 & 0.56 & \textbf{0.86} & 0.00 & 0.02 & 0.00 & 0.18 & 0.02 & 0.00 & 0.19 \\
\modelname (D) 
& 0.34 & 0.00 & 0.56 & 0.22 & 0.00 & 0.00 & 0.00 & 0.00 & 0.00 & 0.00 & 0.11 \\
\modelname (T) 
& \textbf{0.90} & \textbf{0.12} & \textbf{0.82} & 0.84 & \textbf{0.08} & \textbf{0.28} & \textbf{0.10} & \textbf{0.76} & \textbf{0.08} & 0.00 & \textbf{0.40} \\
\bottomrule
\end{tabular}}
\end{subtable}
\hfill
\begin{subtable}[t]{0.44\textwidth}
\centering
\resizebox{\linewidth}{!}{
\begin{tabular}{l|cccccccccc|c}
\toprule
Episodes & T1 & T2 & T3 & T4 & T5 & T6 & T7 & T8 & T9 & T10 & Avg \\
\midrule
$|S|=1$  & 0.88 & 0.00 & \textbf{0.82} & \textbf{0.74} & 0.00 & 0.08 & 0.00 & \textbf{0.72} & \textbf{0.04} & 0.00 & \textbf{0.33} \\
$|S|=3$  & 0.94 & 0.00 & \textbf{0.82} & 0.54 & 0.06 & 0.20 & 0.00 & \textbf{0.72} & 0.00 & 0.00 & \textbf{0.33} \\
$|S|=5$  & 0.96 & \textbf{0.02} & 0.78 & 0.34 & \textbf{0.06} & \textbf{0.38} & 0.00 & 0.68 & \textbf{0.04} & 0.00 & \textbf{0.32} \\
$|S|=10$ & \textbf{1.00} & 0.00 & 0.78 & 0.44 & 0.00 & 0.34 & 0.00 & \textbf{0.72} & 0.00 & 0.00 & \textbf{0.33} \\
\bottomrule
\end{tabular}}
\end{subtable}
\caption{\textbf{Ablations and analyses on task-conditioned weight generation.}
(left) Dataset- (D) vs task- (T) oriented conditioning on LIBERO-Spatial.
(right) Effect of support size $|S|$ for constructing task embeddings on LIBERO-Spatial.}
\label{tab:ablation_grid}
\end{table*}

%% file: tables/table_loss_abl.tex
\begin{table}[!h]
\centering

\resizebox{\columnwidth}{!}{
\begin{tabular}{l|cccccccccc|c}
\toprule
Loss & T1 & T2 & T3 & T4 & T5 & T6 & T7 & T8 & T9 & T10 & Avg \\
\midrule
$\mathcal{L}_{MSE}$ & 0.00 & 0.00 & 0.00 & 0.00 & 0.00 & 0.00 & 0.00 & 0.00 & 0.00 & 0.00 & 0.00 \\
$\mathcal{L}_{MSE}, \mathcal{L}_{cos}$  & 0.00 & 0.00 & 0.00 & 0.00 & 0.00 & 0.00 & 0.00 & 0.00 & 0.00 & 0.00 & 0.00 \\
$\mathcal{L}_{MSE}, \mathcal{L}_{scale}$  & 0.78 & 0.00 & 0.50 & \textbf{0.80} & 0.00 & \textbf{0.16} & 0.00 & 0.46 & 0.00 & 0.00 & 0.27 \\
$\mathcal{L}_{MSE}, \mathcal{L}_{scale}, \mathcal{L}_{cos}$ & 0.\textbf{88} & 0.00 & \textbf{0.82} & 0.74 & 0.00 & 0.08 & 0.00 & \textbf{0.72} & \textbf{0.04} & 0.00 & \textbf{0.33} \\

\bottomrule
\end{tabular}
}
\vspace{0.2cm}
\caption{\textbf{Loss ablation.} Success rates per task and average performance on LIBERO-Spatial when removing components of the training objective.}
\label{tab:loss_abl}
\end{table}

%% file: tables/task_based_training.tex
\begin{table}[!h]
\centering

\scriptsize
\setlength{\tabcolsep}{4pt}
\renewcommand{\arraystretch}{1.0}
\resizebox{0.45\linewidth}{!}{
\begin{tabular}{lccc}
\toprule
Test dataset & Task & \modelname & MT-VLA \\
\midrule
LIBERO-Goal    & T5  & 100\% & 98\%  \\
LIBERO-Goal    & T9  & 90\%  & 92\%  \\
LIBERO-Object  & T5  & 95\%  & 100\% \\
LIBERO-Object  & T10 & 98\%  & 84\%  \\
LIBERO-Spatial & T6  & 90\%  & 94\%  \\
LIBERO-Spatial & T8  & 82\%  & 86\%  \\
LIBERO-10      & T1  & 66\%  & 80\%  \\
LIBERO-10      & T7  & 60\%  & 80\%  \\
\bottomrule
\end{tabular}}
\vspace{0.2cm}
\caption{Cross-task generalization within datasets.}
\label{tab:task_based_training}
\end{table}

%% file: tables/speed_memory.tex
\begin{table}[!ht]
\centering
\setlength{\tabcolsep}{3pt}
\renewcommand{\arraystretch}{0.9}
\resizebox{0.5\linewidth}{!}{
\begin{tabular}{lcccc}
\toprule
\textbf{Rank} & \textbf{Params} & \textbf{Fraction} & \textbf{Time} & \textbf{Memory} \\
\midrule
r16 (ours) & 46.1M & 1.27\% & 0.07s & 1.74GB \\
r64 & 184.4M & 5.09\% & 0.20s & 3.81GB \\
r256 & 737.5M & 20.37\% & 0.71s & 12.09GB \\
r512 & 1.48B & 40.75\% & 1.42s & 23.5GB \\
\midrule
r1257 ($\pi_{0.5}$) & 3.62B & 100.00\% & -- & \textbf{Out Of Mem.} \\
\bottomrule
\end{tabular}}
\vspace{0.2cm}
\caption{Scaling cost.}
\label{tab:rank_scaling_cost}
\end{table}

%% file: figures/app_qualitative_spatial.tex
\begin{figure*}[h]
    \centering


    \begin{subfigure}[b]{.98\linewidth}
        \includegraphics[width=\linewidth]{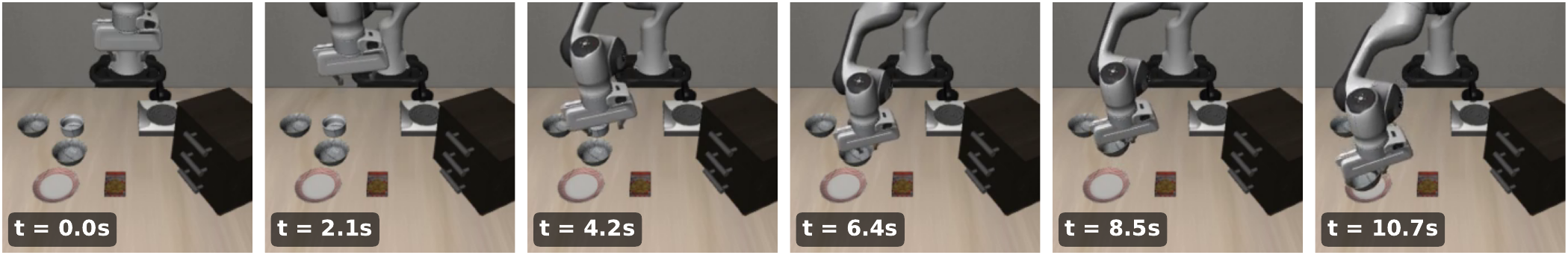}
        \caption{Task 1: \textit{pick up the black bowl between the plate and the ramekin and place it on the plate.}}
    \end{subfigure}

    \vspace{2mm}

    \begin{subfigure}[b]{.98\linewidth}
        \includegraphics[width=\linewidth]{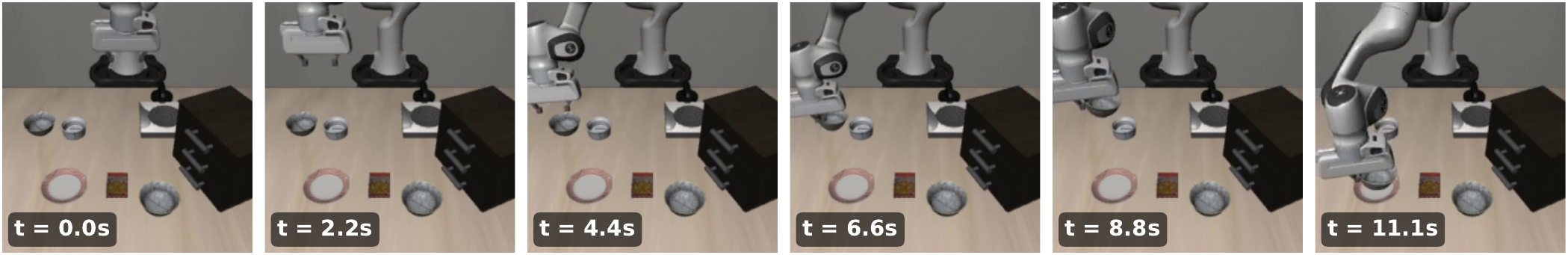}
        \caption{Task 2: \textit{.pick up the black bowl next to the ramekin and place it on the plate}}
    \end{subfigure}

    \vspace{2mm}

    \begin{subfigure}[b]{.98\linewidth}
        \includegraphics[width=\linewidth]{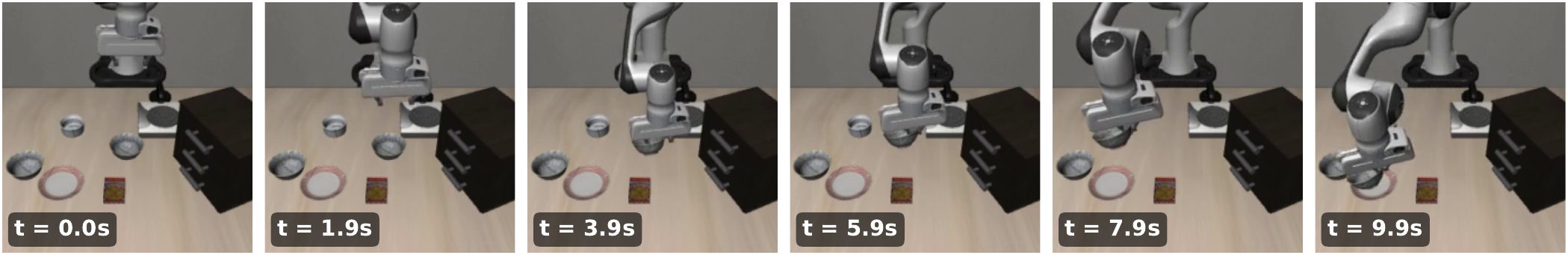}
        \caption{Task 3: \textit{pick up the black bowl from table center and place it on the plate.}}
    \end{subfigure}

    \vspace{2mm}

    \begin{subfigure}[b]{.98\linewidth}
        \includegraphics[width=\linewidth]{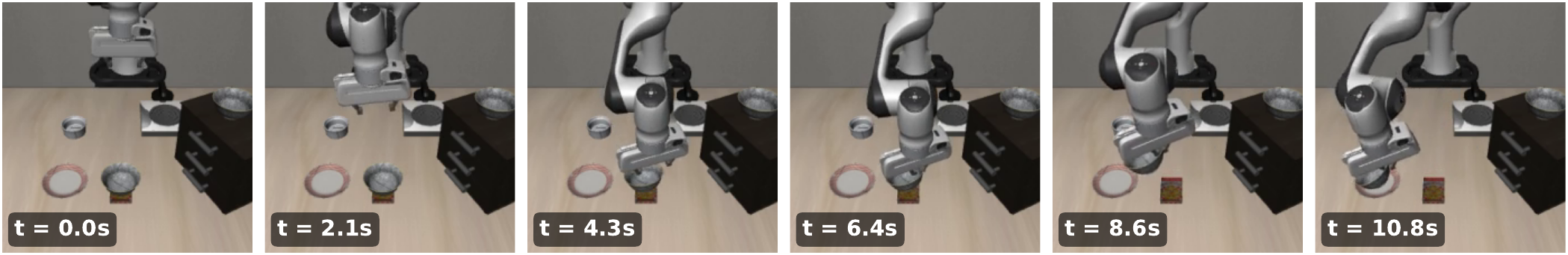}
        \caption{Task 4: \textit{pick up the black bowl on the cookie box and place it on the plate.}}
    \end{subfigure}

    \vspace{2mm}

    \begin{subfigure}[b]{.98\linewidth}
        \includegraphics[width=\linewidth]{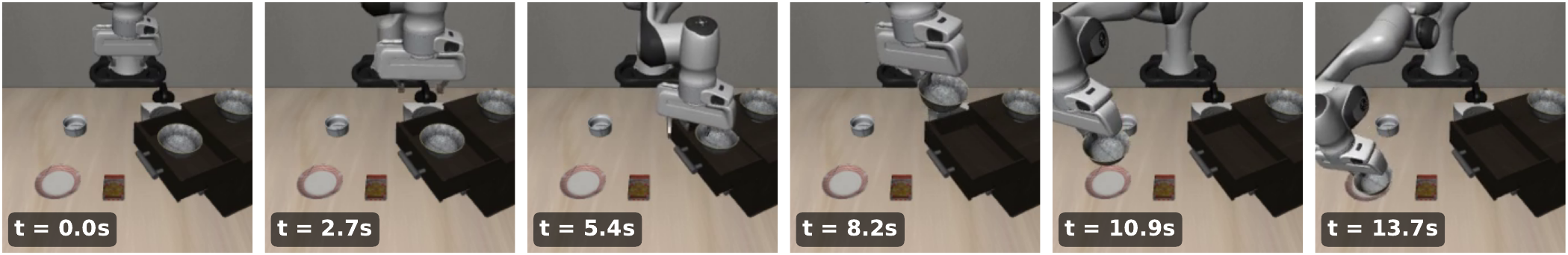}
        \caption{Task 5: \textit{pick up the black bowl in the top drawer of the cabinet and place it on the plate.}}
    \end{subfigure}

    \vspace{2mm}

    \begin{subfigure}[b]{.98\linewidth}
        \includegraphics[width=\linewidth]{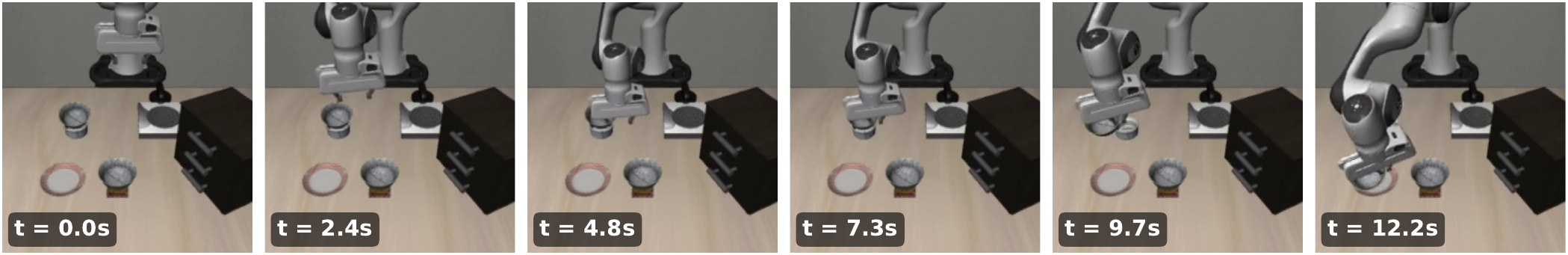}
        \caption{Task 6: \textit{pick up the black bowl on the ramekin and place it on the plate.}}
    \end{subfigure}

    \vspace{2mm}
    



    
    \caption{\textbf{Additional zero-shot qualitative rollouts on LIBERO-Spatial.}}
    \label{fig:add_qualitative_spatial}
\end{figure*}

%% file: figures/app_qualitative_spatial2.tex
\begin{figure*}[h]
    \centering
    
    \begin{subfigure}[b]{.98\linewidth}
        \includegraphics[width=\linewidth]{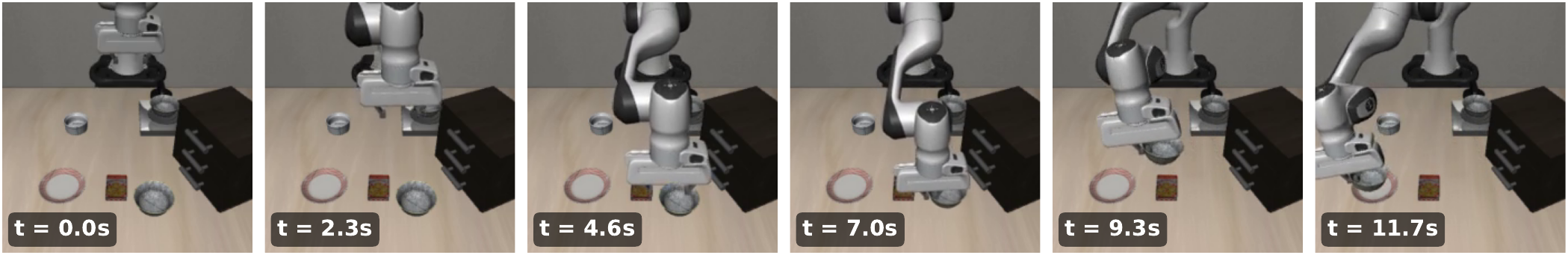}
        \caption{Task 7: \textit{pick up the black bowl next to the cookie box and place it on the plate.}}
    \end{subfigure}

    \vspace{2mm}

    \begin{subfigure}[b]{.98\linewidth}
        \includegraphics[width=\linewidth]{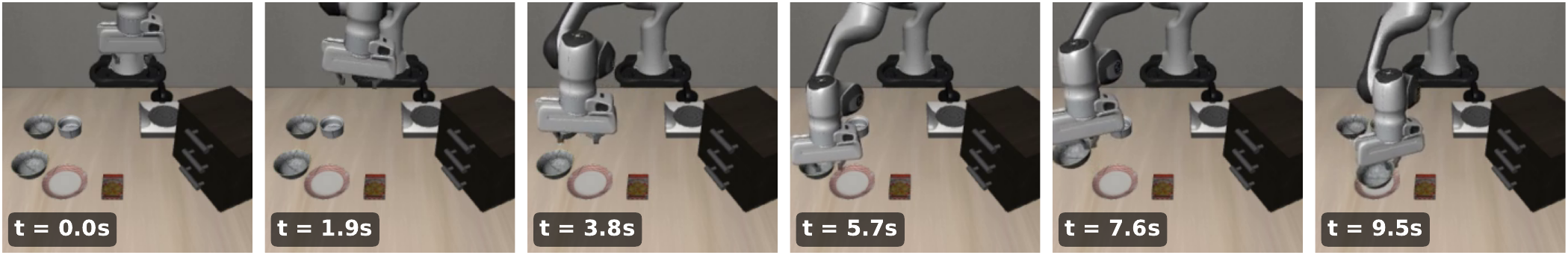}
        \caption{Task 9: \textit{pick up the black bowl next to the plate and place it on the plate.}}
    \end{subfigure}

    \vspace{2mm}
    
    \caption{\textbf{Additional zero-shot qualitative rollouts on LIBERO-Spatial.}}
    \label{fig:add_qualitative_spatial2}
\end{figure*}

%% file: figures/app_qualitative_object.tex
\begin{figure*}[h]
    \centering

    
    \begin{subfigure}[b]{.98\linewidth}
        \includegraphics[width=\linewidth]{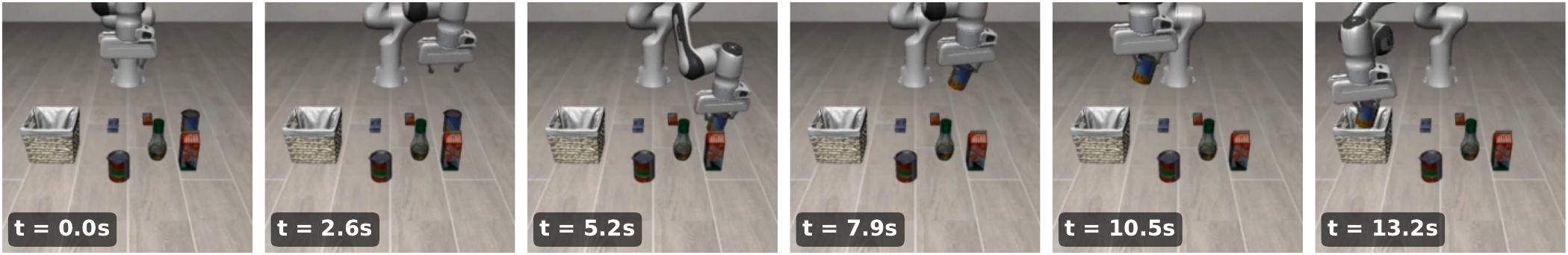}
        \caption{Task 1: \textit{pick up the alphabet soup and place it in the basket.}}
    \end{subfigure}
    
    \vspace{2mm} 
    
    \begin{subfigure}[b]{.98\linewidth}
        \includegraphics[width=\linewidth]{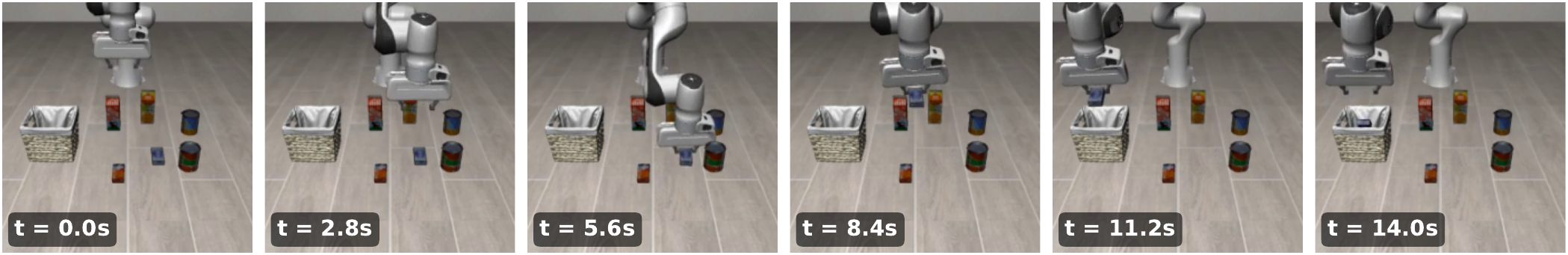}
        \caption{Task 2: \textit{pick up the cream cheese and place it in the basket.}}
    \end{subfigure}

    \vspace{2mm}
    
    \begin{subfigure}[b]{.98\linewidth}
        \includegraphics[width=\linewidth]{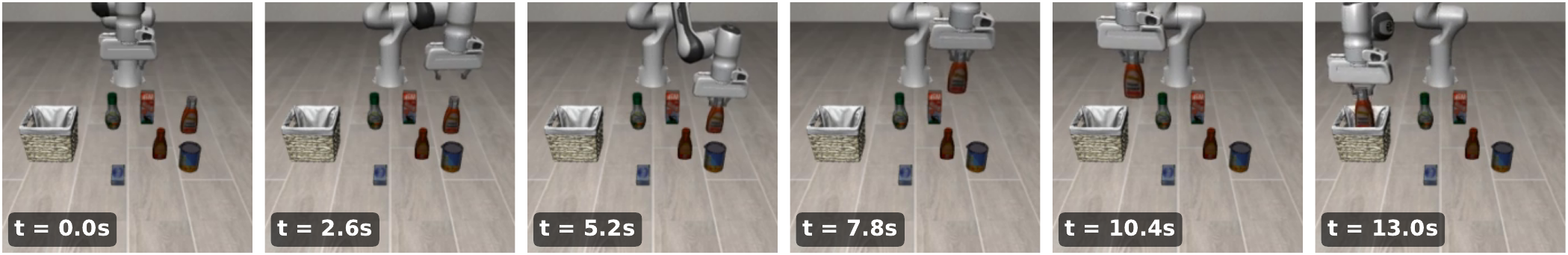}
        \caption{Task 5: \textit{pick up the ketchup and place it in the basket.}}
    \end{subfigure}

    \vspace{2mm} 

    \caption{\textbf{Additional zero-shot qualitative rollouts on LIBERO-Object.}}
    \label{fig:add_qualitative_object}
\end{figure*}

%% file: figures/app_qualitative_goal.tex
\begin{figure*}[!ht]
    \centering


    \begin{subfigure}[b]{.98\linewidth}
        \includegraphics[width=\linewidth]{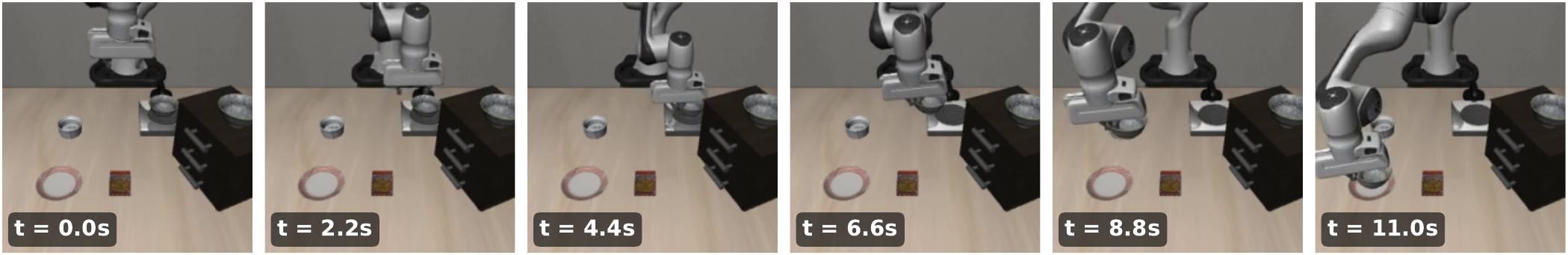}
        \caption{Task 2: \textit{put the bowl on the stove.}}
    \end{subfigure}

    \vspace{2mm}

    \begin{subfigure}[b]{.98\linewidth}
        \includegraphics[width=\linewidth]{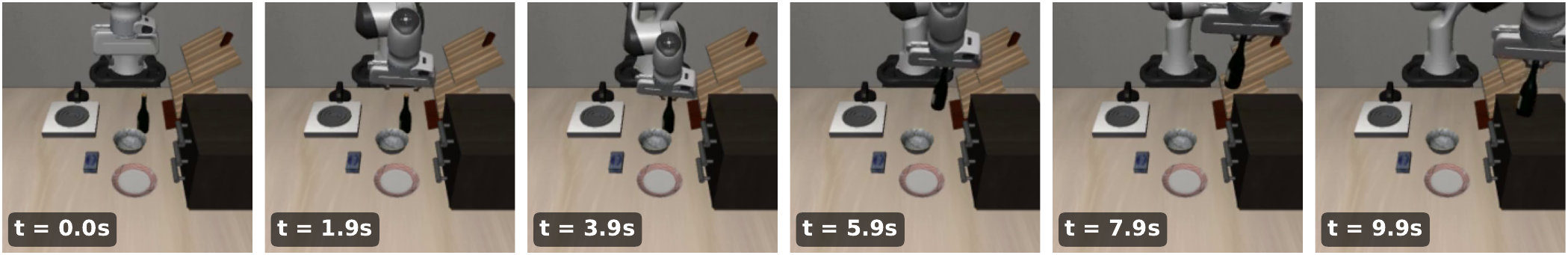}
        \caption{Task 3: \textit{put the wine bottle on top of the cabinet.}}
    \end{subfigure}

    \vspace{2mm}

    \begin{subfigure}[b]{.98\linewidth}
        \includegraphics[width=\linewidth]{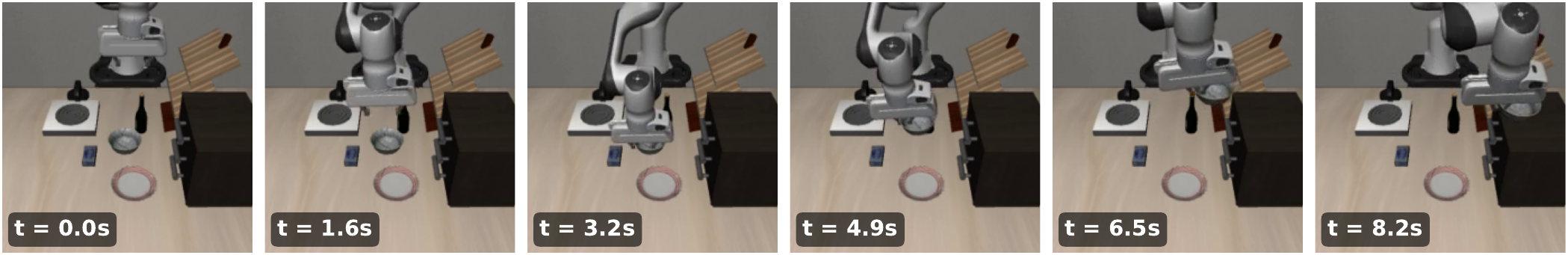}
        \caption{Task 5: \textit{put the bowl on top of the cabinet.}}
    \end{subfigure}

    \vspace{2mm}
    
    \begin{subfigure}[b]{.98\linewidth}
        \includegraphics[width=\linewidth]{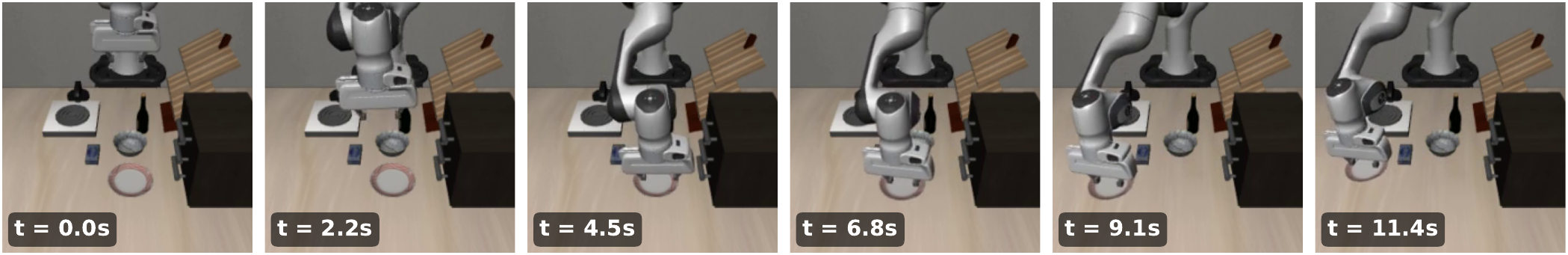}
        \caption{Task 6: \textit{push the plate to the front of the stove.}}
    \end{subfigure}
    
    \vspace{2mm}

    \begin{subfigure}[b]{.98\linewidth}
        \includegraphics[width=\linewidth]{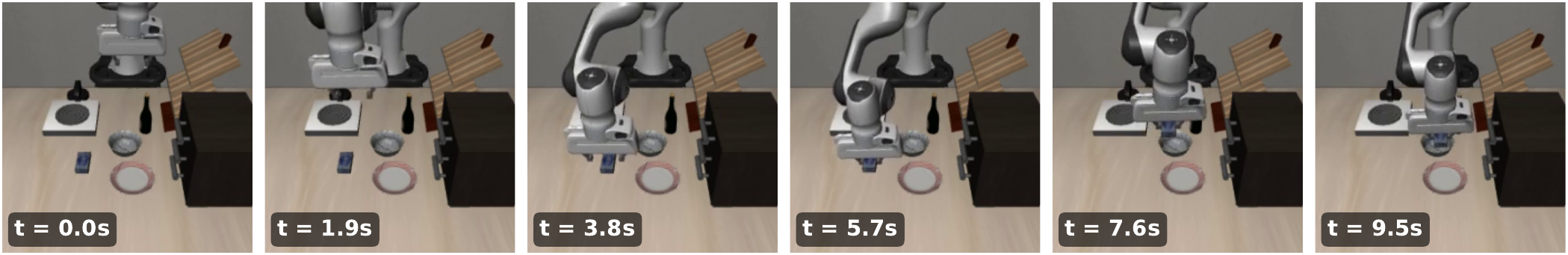}
        \caption{Task 7: \textit{put the cream cheese in the bowl.}}
    \end{subfigure}

    \vspace{2mm}

    \begin{subfigure}[b]{.98\linewidth}
        \includegraphics[width=\linewidth]{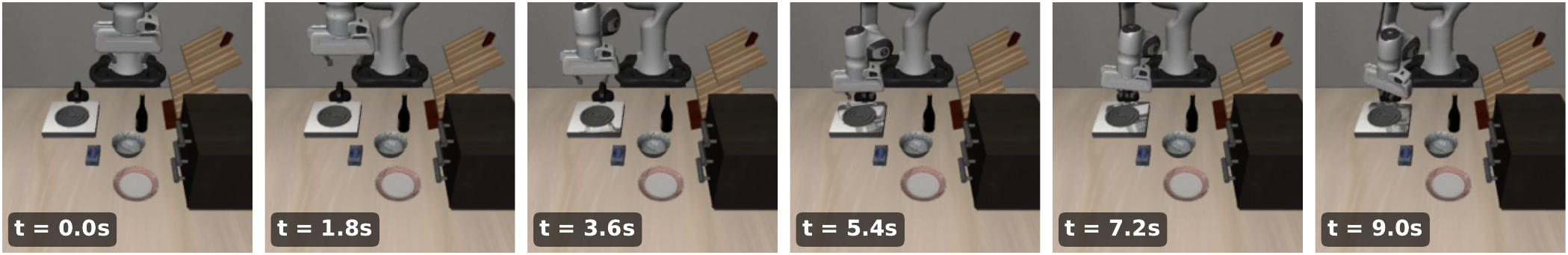}
       \caption{Task 8: \textit{turn on the stove.}}
    \end{subfigure}

    \vspace{2mm}

    \begin{subfigure}[b]{.98\linewidth}
        \includegraphics[width=\linewidth]{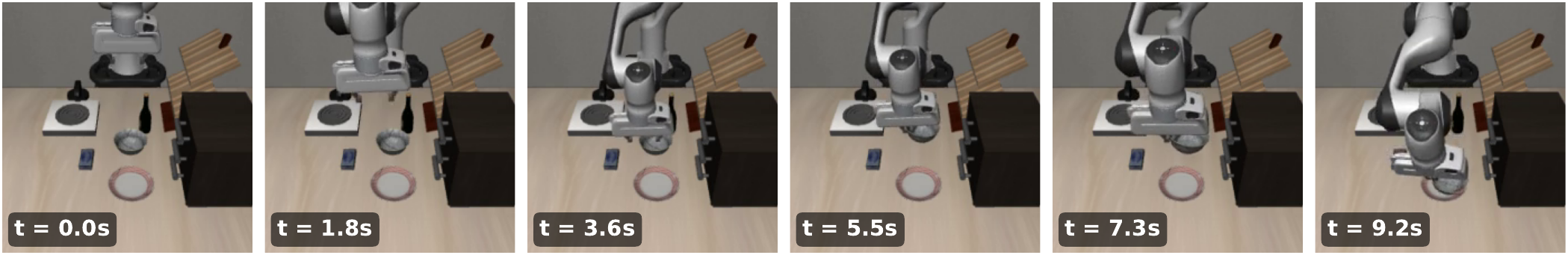}
        \caption{Task 9: \textit{put the bowl on the plate.}}
    \end{subfigure}

    \vspace{2mm}
    
    \caption{\textbf{Additional zero-shot qualitative rollouts on LIBERO-Goal.}}
    \label{fig:add_qualitative_goal}
\end{figure*}

%% file: figures/app_qualitative_10.tex
\begin{figure*}[h]
    \centering

    \begin{subfigure}[b]{.98\linewidth}
        \includegraphics[width=\linewidth]{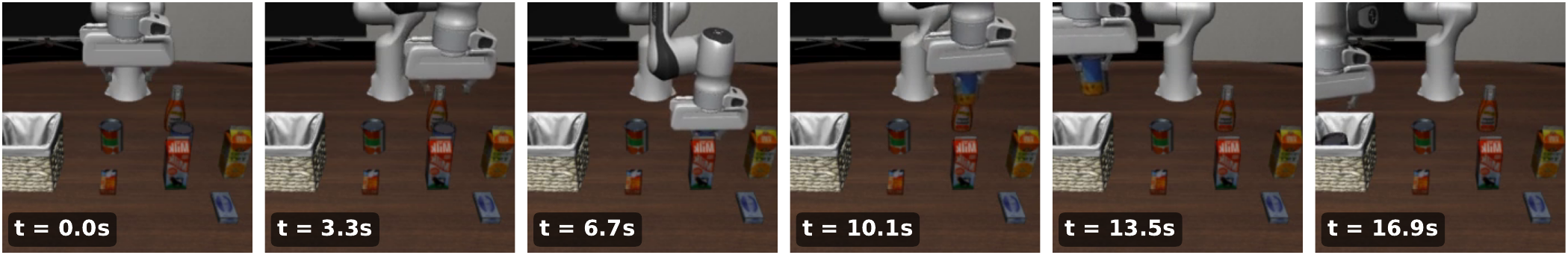}
        \caption{Task 1A: \textit{put the alphabet soup in the basket.}}
    \end{subfigure}

    \vspace{2mm}

    \begin{subfigure}[b]{.98\linewidth}
        \includegraphics[width=\linewidth]{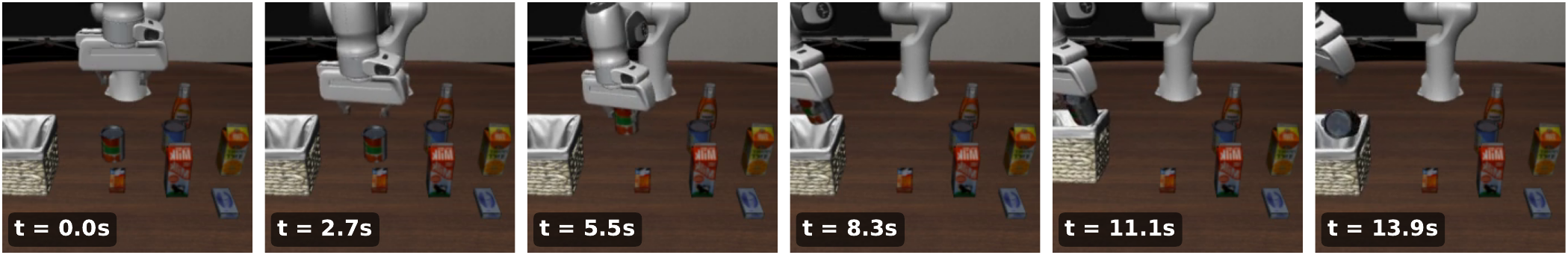}
        \caption{Task 1B: \textit{put the tomato sauce in the basket.}}
    \end{subfigure}

    \vspace{2mm}

    \begin{subfigure}[b]{.98\linewidth}
        \includegraphics[width=\linewidth]{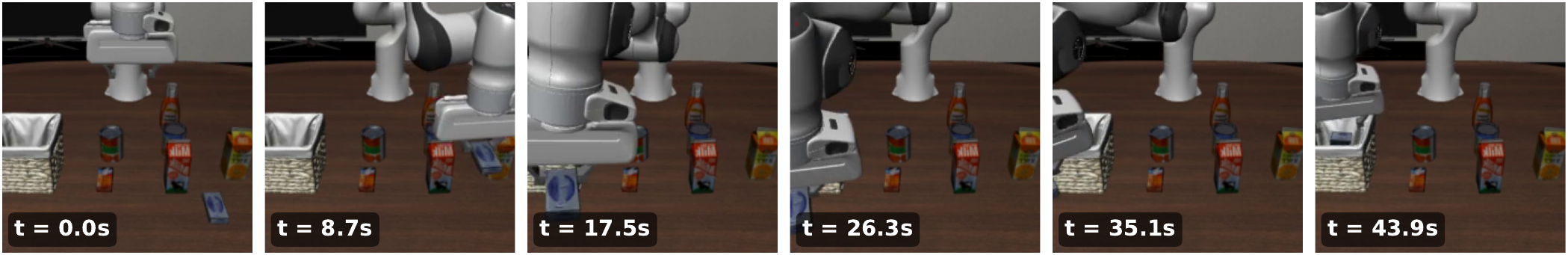}
        \caption{Task 2A: \textit{put the cream cheese box in the basket.}}
    \end{subfigure}

    \vspace{2mm}

    \begin{subfigure}[b]{.98\linewidth}
        \includegraphics[width=\linewidth]{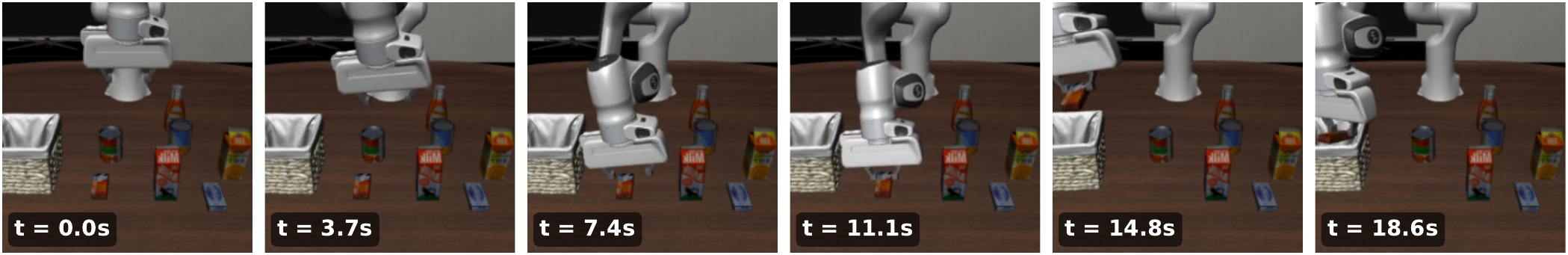}
        \caption{Task 2B: \textit{put butter in the basket.}}
    \end{subfigure}

    \vspace{2mm}
    
    \begin{subfigure}[b]{.98\linewidth}
        \includegraphics[width=\linewidth]{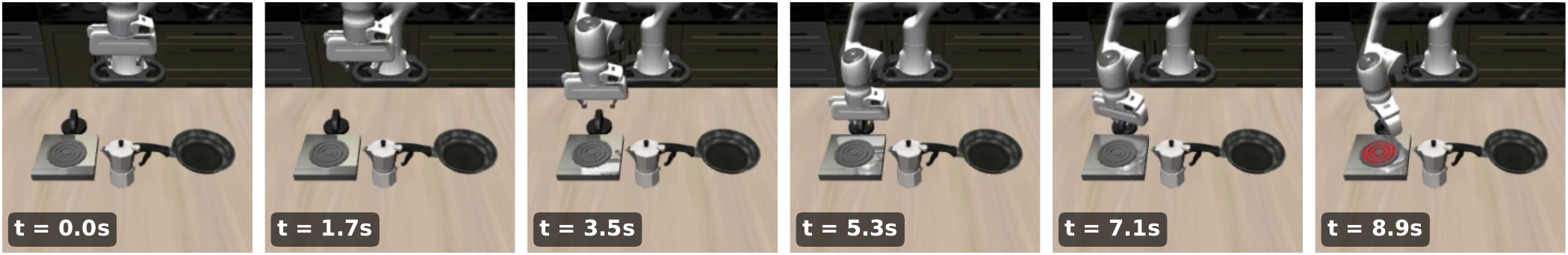}
        \caption{Task 3A: \textit{turn on the stove.}}
    \end{subfigure}

    \vspace{2mm}

    \begin{subfigure}[b]{.98\linewidth}
        \includegraphics[width=\linewidth]{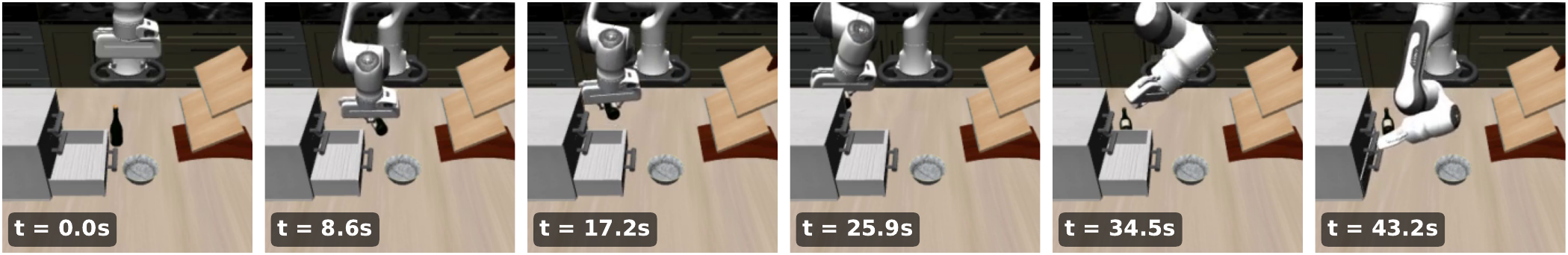}
        \caption{Task 4A: \textit{close the bottom drawer of the cabinet.}}
    \end{subfigure}

    \vspace{2mm}

    \begin{subfigure}[b]{.98\linewidth}
        \includegraphics[width=\linewidth]{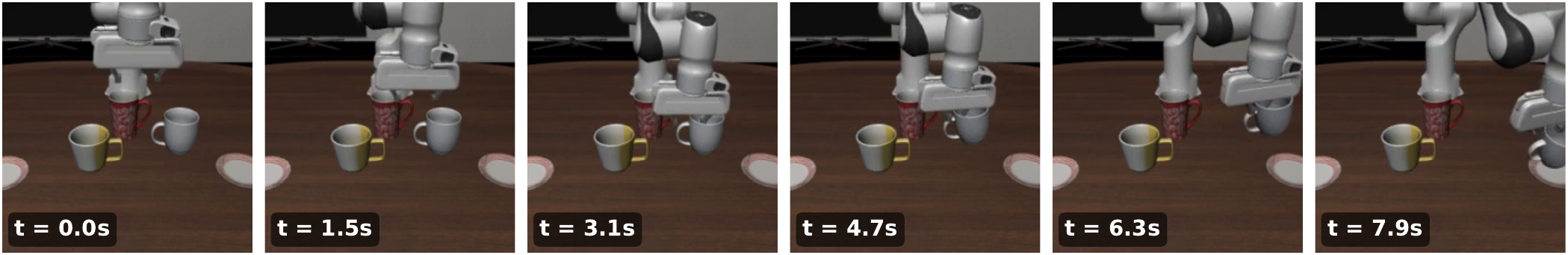}
        \caption{Task 5A: \textit{put the white mug on the left plate.}}
    \end{subfigure}

    \vspace{2mm}

    \caption{\textbf{Additional zero-shot qualitative rollouts on LIBERO-10.}}
    \label{fig:add_qualitative_10}
\end{figure*}

%% file: figures/app_qualitative_102.tex
\begin{figure*}[h]
    \centering

    \begin{subfigure}[b]{.98\linewidth}
        \includegraphics[width=\linewidth]{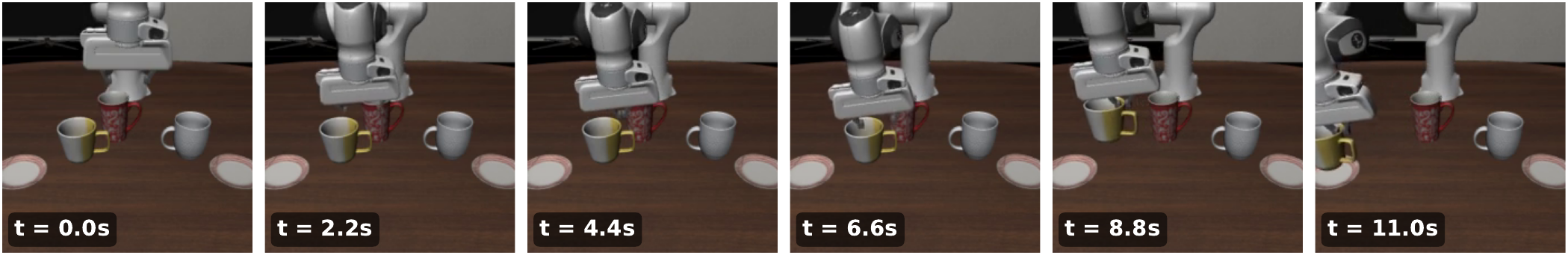}
        \caption{Task 5B: \textit{put the yellow and white mug on the right plate.}}
    \end{subfigure}

    \vspace{2mm}

    \begin{subfigure}[b]{.98\linewidth}
        \includegraphics[width=\linewidth]{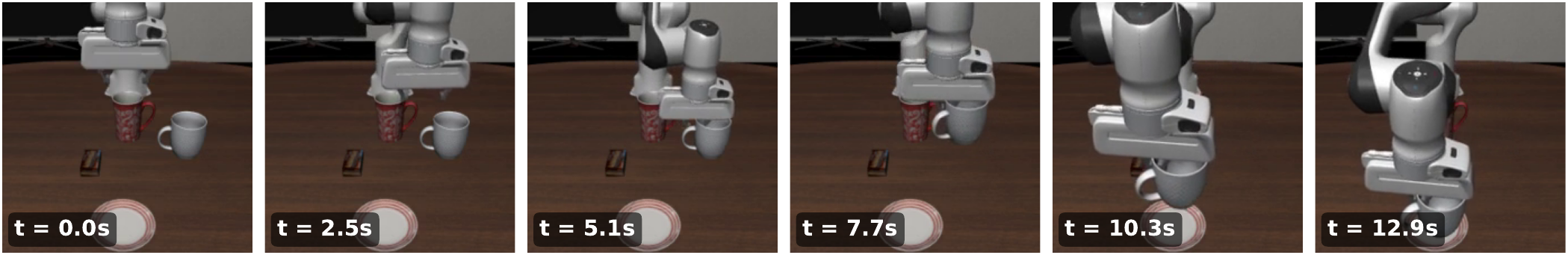}
        \caption{Task 7A: \textit{put the white mug on the plate.}}
    \end{subfigure}

    \vspace{2mm}

    \begin{subfigure}[b]{.98\linewidth}
        \includegraphics[width=\linewidth]{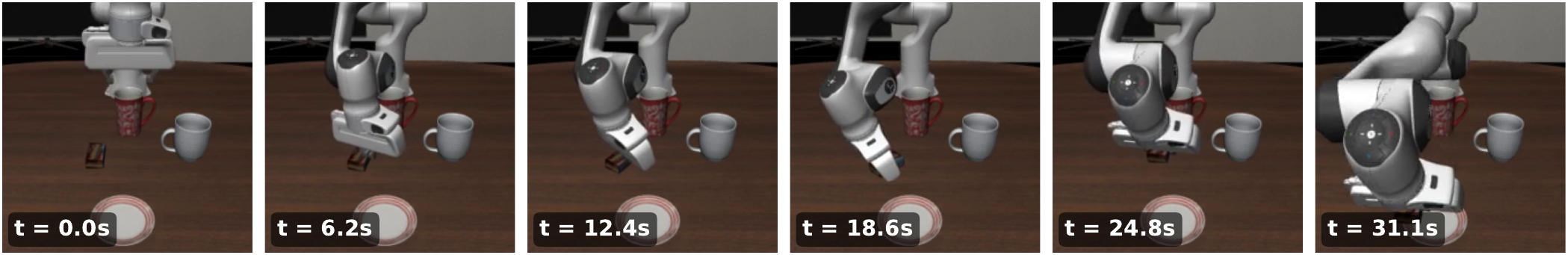}
        \caption{Task 7B: \textit{put the chocolate pudding to the right of the plate.}}
    \end{subfigure}

    \vspace{2mm}

    \begin{subfigure}[b]{.98\linewidth}
        \includegraphics[width=\linewidth]{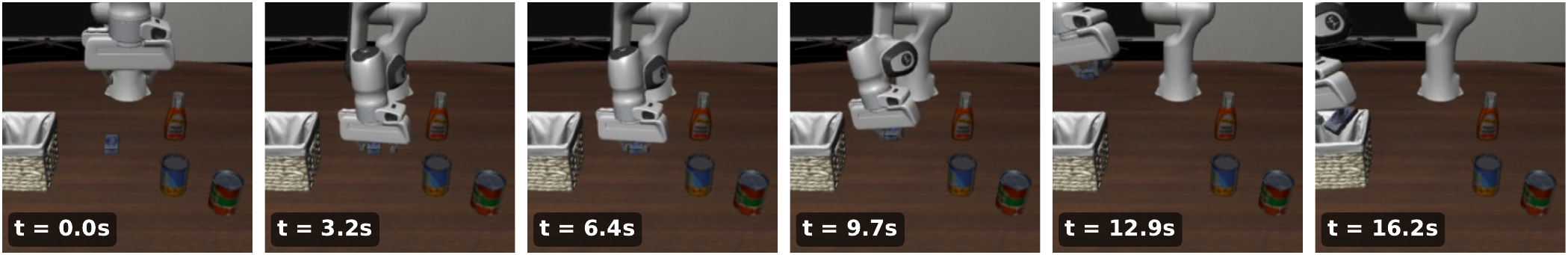}
        \caption{Task 8B: \textit{put the cream cheese box in the basket.}}
    \end{subfigure}

    \vspace{2mm}

    \begin{subfigure}[b]{.98\linewidth}
        \includegraphics[width=\linewidth]{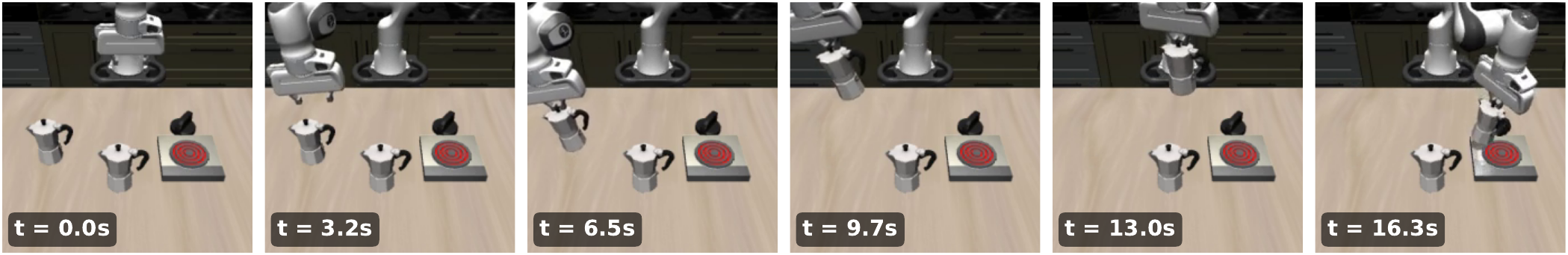}
        \caption{Task 9A: \textit{put the rightmost moka pot on the stove.}}
    \end{subfigure}

    \vspace{2mm}
    \begin{subfigure}[b]{.98\linewidth}
        \includegraphics[width=\linewidth]{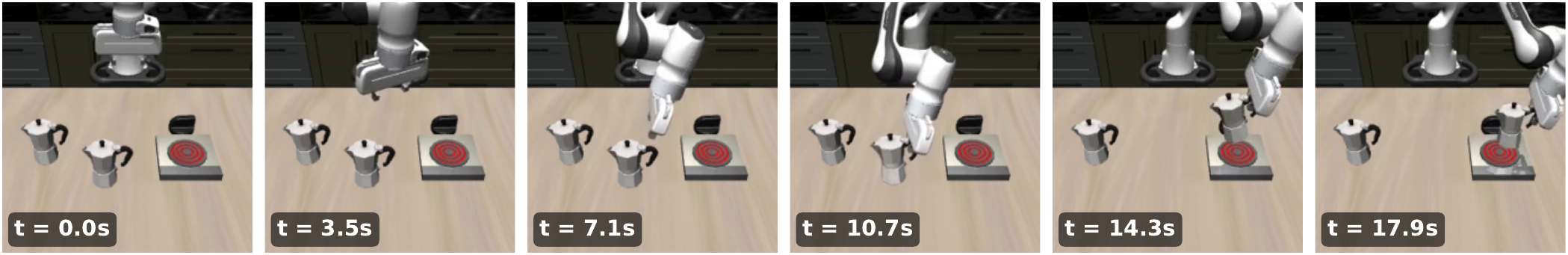}
        \caption{Task 9B: \textit{put the leftmost moka pot on the stove.}}
    \end{subfigure}

    \vspace{2mm}

    \begin{subfigure}[b]{.98\linewidth}
        \includegraphics[width=\linewidth]{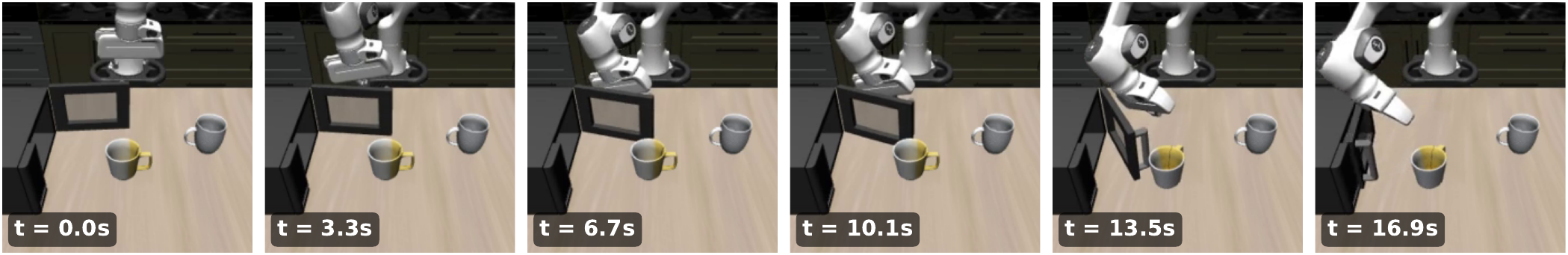}
        \caption{Task 10B: \textit{close the microwave.}}
    \end{subfigure}

    \vspace{2mm}

    \caption{\textbf{Additional zero-shot qualitative rollouts on LIBERO-10.}}
    \label{fig:add_qualitative_102}
\end{figure*}

%% file: figures/app_qualitative_real.tex
\begin{figure*}[h]
    \centering

    \begin{subfigure}[b]{.98\linewidth}
        \includegraphics[width=\linewidth]{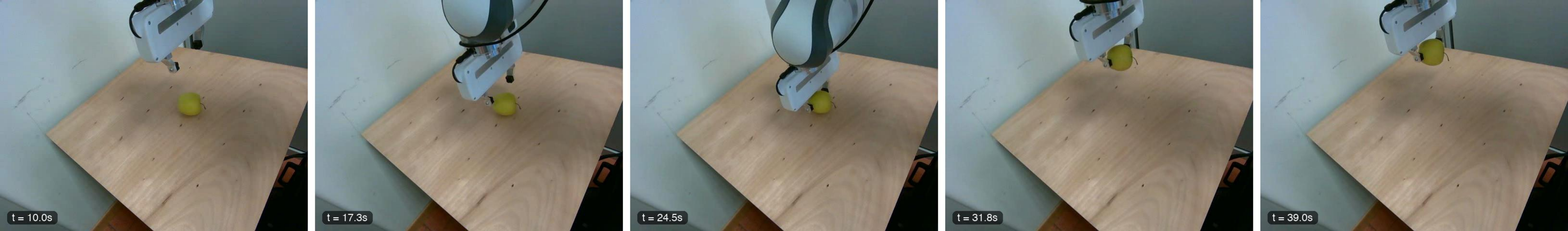}
        \caption{Task 1: \textit{pick up the apple.}}
    \end{subfigure}

    \vspace{2mm}

    \begin{subfigure}[b]{.98\linewidth}
        \includegraphics[width=\linewidth]{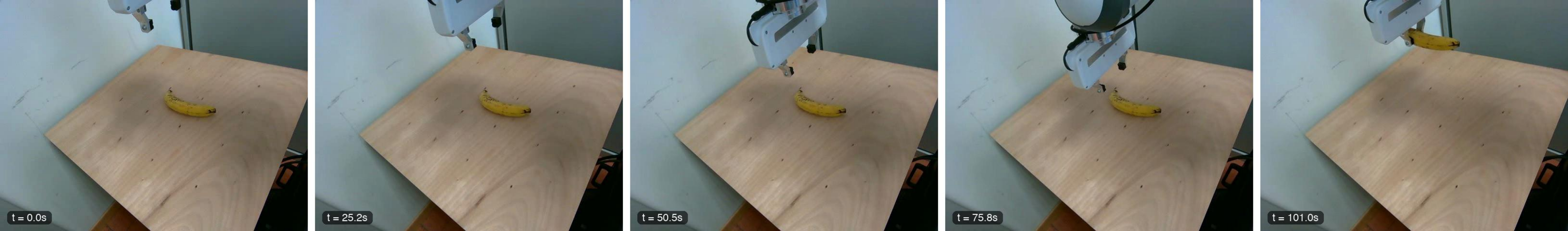}
        \caption{Task 2: \textit{pick up the banana.}}
    \end{subfigure}

    \vspace{2mm}

    \begin{subfigure}[b]{.98\linewidth}
        \includegraphics[width=\linewidth]{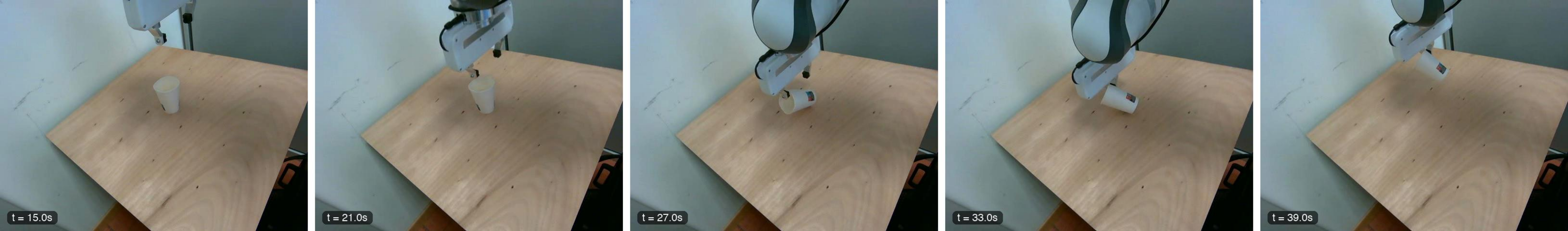}
        \caption{Task 3: \textit{pick up the cup.}}
    \end{subfigure}

    \vspace{2mm}

    \caption{\textbf{Additional qualitative rollouts in Real-World.}}
    \label{fig:add_qualitative_real}
\end{figure*}

%% file: tables/task_table.tex
\begin{itemize}
    \item \textbf{Task 1:} \texttt{pick up the black bowl between the plate and the ramekin and place it on the plate}
    \item \textbf{Task 2:} \texttt{pick up the black bowl next to the ramekin and place it on the plate}
    \item \textbf{Task 3:} \texttt{pick up the black bowl from table center and place it on the plate}
    \item \textbf{Task 4:} \texttt{pick up the black bowl on the cookie box and place it on the plate}
    \item \textbf{Task 5:} \texttt{pick up the black bowl in the top drawer of the wooden\\cabinet and place it on the plate}
    \item \textbf{Task 6:} \texttt{pick up the black bowl on the ramekin and place it on the plate}
    \item \textbf{Task 7:} \texttt{pick up the black bowl next to the cookie box and place it on the plate}
    \item \textbf{Task 8:} \texttt{pick up the black bowl on the stove and place it on the plate}
    \item \textbf{Task 9:} \texttt{pick up the black bowl next to the plate and place it on the plate}
    \item \textbf{Task 10:} \texttt{pick up the black bowl on the wooden cabinet and place it on the plate}
\end{itemize}

\noindent \textbf{LIBERO-Object}
This dataset isolates visual object grounding. The kinematic intent (\texttt{`pick up... and place it in the basket'}) is invariant, serving as a highly stable, dense prior for basic manipulation within the meta-learned manifold.

\begin{itemize}
    \item \textbf{Task 1:} \texttt{pick up the alphabet soup and place it in the basket}
    \item \textbf{Task 2:} \texttt{pick up the cream cheese and place it in the basket}
    \item \textbf{Task 3:} \texttt{pick up the salad dressing and place it in the basket}
    \item \textbf{Task 4:} \texttt{pick up the bbq sauce and place it in the basket}
    \item \textbf{Task 5:} \texttt{pick up the ketchup and place it in the basket}
    \item \textbf{Task 6:} \texttt{pick up the tomato sauce and place it in the basket}
    \item \textbf{Task 7:} \texttt{pick up the butter and place it in the basket}
    \item \textbf{Task 8:} \texttt{pick up the milk and place it in the basket}
    \item \textbf{Task 9:} \texttt{pick up the chocolate pudding and place it in the basket}
    \item \textbf{Task 10:} \texttt{pick up the orange juice and place it in the basket}
\end{itemize}

\noindent \textbf{LIBERO-Goal}
This dataset tests goal-directed interactions. Tasks containing the \texttt{`open'} primitive act as semantic boundaries for zero-shot synthesis when held out of the training distribution.

\begin{itemize}
    \item \textbf{Task 1:} \texttt{open the middle drawer of the cabinet}
    \item \textbf{Task 2:} \texttt{put the bowl on the stove}
    \item \textbf{Task 3:} \texttt{put the wine bottle on top of the cabinet}
    \item \textbf{Task 4:} \texttt{open the top drawer and put the bowl inside}
    \item \textbf{Task 5:} \texttt{put the bowl on top of the cabinet}
    \item \textbf{Task 6:} \texttt{push the plate to the front of the stove}
    \item \textbf{Task 7:} \texttt{put the cream cheese in the bowl}
    \item \textbf{Task 8:} \texttt{turn on the stove}
    \item \textbf{Task 9:} \texttt{put the bowl on the plate}
    \item \textbf{Task 10:} \texttt{put the wine bottle on the rack}\\
\end{itemize}

\noindent \textbf{LIBERO-10 (Long-Horizon Composition)}
This dataset introduces severe temporal and compositional complexity. The prompts demand the sequential execution of multiple distinct primitives, testing the temporal consistency of the generated LoRA parameters.

\begin{itemize}
    \item \textbf{Task 1:} \texttt{LIVING ROOM SCENE2 put both the alphabet soup and the\\tomato sauce in the basket}
    \item \textbf{Task 2:} \texttt{LIVING ROOM SCENE2 put both the cream cheese box and the butter in the basket}
    \item \textbf{Task 3:} \texttt{KITCHEN SCENE3 turn on the stove and put the moka pot on it}
    \item \textbf{Task 4:} \texttt{KITCHEN SCENE4 put the black bowl in the bottom drawer\\of the cabinet and close it}
    \item \textbf{Task 5:} \texttt{LIVING ROOM SCENE5 put the white mug on the left plate\\and put the yellow and white mug on the right plate}
    \item \textbf{Task 6:} \texttt{STUDY SCENE1 pick up the book and place it in the back\\compartment of the caddy}
    \item \textbf{Task 7:} \texttt{LIVING ROOM SCENE6 put the white mug on the plate and\\put the chocolate pudding to the right of the plate}
    \item \textbf{Task 8:} \texttt{LIVING ROOM SCENE1 put both the alphabet soup and the\\cream cheese box in the basket}
    \item \textbf{Task 9:} \texttt{KITCHEN SCENE8 put both moka pots on the stove}
    \item \textbf{Task 10:} \texttt{KITCHEN SCENE6 put the yellow and white mug in the\\microwave and close it}
\end{itemize}

%% file: main.bib
@misc{droid_2024,
    title={DROID: A Large-Scale In-The-Wild Robot Manipulation Dataset},
    author  = {Alexander Khazatsky and Karl Pertsch and Suraj Nair and Ashwin Balakrishna and Sudeep Dasari and Siddharth Karamcheti and Soroush Nasiriany and Mohan Kumar Srirama and Lawrence Yunliang Chen and Kirsty Ellis and Peter David Fagan and Joey Hejna and Masha Itkina and Marion Lepert and Yecheng Jason Ma and Patrick Tree Miller and Jimmy Wu and Suneel Belkhale and Shivin Dass and Huy Ha and Arhan Jain and Abraham Lee and Youngwoon Lee and Marius Memmel and Sungjae Park and Ilija Radosavovic and Kaiyuan Wang and Albert Zhan and Kevin Black and Cheng Chi and Kyle Beltran Hatch and Shan Lin and Jingpei Lu and Jean Mercat and Abdul Rehman and Pannag R Sanketi and Archit Sharma and Cody Simpson and Quan Vuong and Homer Rich Walke and Blake Wulfe and Ted Xiao and Jonathan Heewon Yang and Arefeh Yavary and Tony Z. Zhao and Christopher Agia and Rohan Baijal and Mateo Guaman Castro and Daphne Chen and Qiuyu Chen and Trinity Chung and Jaimyn Drake and Ethan Paul Foster and Jensen Gao and Vitor Guizilini and David Antonio Herrera and Minho Heo and Kyle Hsu and Jiaheng Hu and Muhammad Zubair Irshad and Donovon Jackson and Charlotte Le and Yunshuang Li and Kevin Lin and Roy Lin and Zehan Ma and Abhiram Maddukuri and Suvir Mirchandani and Daniel Morton and Tony Nguyen and Abigail O'Neill and Rosario Scalise and Derick Seale and Victor Son and Stephen Tian and Emi Tran and Andrew E. Wang and Yilin Wu and Annie Xie and Jingyun Yang and Patrick Yin and Yunchu Zhang and Osbert Bastani and Glen Berseth and Jeannette Bohg and Ken Goldberg and Abhinav Gupta and Abhishek Gupta and Dinesh Jayaraman and Joseph J Lim and Jitendra Malik and Roberto Martín-Martín and Subramanian Ramamoorthy and Dorsa Sadigh and Shuran Song and Jiajun Wu and Michael C. Yip and Yuke Zhu and Thomas Kollar and Sergey Levine and Chelsea Finn},
    booktitle = {Robotics: Science and Systems (RSS)},
    year = {2024},
}

@inproceedings{brohan2023rt1,
  title={RT-1: Robotics Transformer for Real-World Control at Scale},
    author={Anthony	Brohan and  Noah Brown and  Justice Carbajal and  Yevgen Chebotar and  Joseph Dabis and  Chelsea Finn and  Keerthana Gopalakrishnan and  Karol Hausman and  Alex Herzog and  Jasmine Hsu and  Julian Ibarz and  Brian Ichter and  Alex Irpan and  Tomas Jackson and  Sally Jesmonth and  Nikhil Joshi and  Ryan Julian and  Dmitry Kalashnikov and  Yuheng Kuang and  Isabel Leal and  Kuang-Huei Lee and  Sergey Levine and  Yao Lu and  Utsav Malla and  Deeksha Manjunath and  Igor Mordatch and  Ofir Nachum and  Carolina Parada and  Jodilyn Peralta and  Emily Perez and  Karl Pertsch and  Jornell Quiambao and  Kanishka Rao and  Michael Ryoo and  Grecia Salazar and  Pannag Sanketi and  Kevin Sayed and  Jaspiar Singh and  Sumedh Sontakke and  Austin Stone and  Clayton Tan and  Huong Tran and  Vincent Vanhoucke and Steve Vega and  Quan Vuong and  Fei Xia and  Ted Xiao and  Peng Xu and  Sichun Xu and  Tianhe Yu and  Brianna Zitkovich},
    booktitle={arXiv preprint arXiv:2212.06817},
    year={2022}
}

@misc{kalashnikov2021mtoptcontinuousmultitaskrobotic,
      title={MT-Opt: Continuous Multi-Task Robotic Reinforcement Learning at Scale}, 
      author={Dmitry Kalashnikov and Jacob Varley and Yevgen Chebotar and Benjamin Swanson and Rico Jonschkowski and Chelsea Finn and Sergey Levine and Karol Hausman},
      year={2021},
      eprint={2104.08212},
      archivePrefix={arXiv},
      primaryClass={cs.RO},
      url={https://arxiv.org/abs/2104.08212}, 
}

@inproceedings{kumar2021rma,
title={Rma: Rapid motor adaptation for legged robots},
author={Kumar, Ashish and Fu, Zipeng and Pathak, Deepak and Malik, Jitendra},
journal={Robotics: Science and Systems},
year={2021}
}

@inproceedings{driess2023palmeembodiedmultimodallanguage,
  title={PaLM-E: An Embodied Multimodal Language Model},
  author={Danny Driess and F. Xia and Mehdi S. M. Sajjadi and Corey Lynch and Aakanksha Chowdhery and Brian Ichter and Ayzaan Wahid and Jonathan Tompson and Quan Ho Vuong and Tianhe Yu and Wenlong Huang and Yevgen Chebotar and Pierre Sermanet and Daniel Duckworth and Sergey Levine and Vincent Vanhoucke and Karol Hausman and Marc Toussaint and Klaus Greff and Andy Zeng and Igor Mordatch and Peter R. Florence},
  booktitle={International Conference on Machine Learning},
  year={2023},
  url={https://api.semanticscholar.org/CorpusID:257364842}
}

@article{hu2022lora,
  title={LoRA: Low-Rank Adaptation of Large Language Models},
  author={J. Edward Hu and Yelong Shen and Phillip Wallis and Zeyuan Allen-Zhu and Yuanzhi Li and Shean Wang and Weizhu Chen},
  journal={ArXiv},
  year={2021},
  volume={abs/2106.09685},
  url={https://api.semanticscholar.org/CorpusID:235458009}
}

@misc{liu2023liberobenchmarkingknowledgetransfer,
      title={LIBERO: Benchmarking Knowledge Transfer for Lifelong Robot Learning}, 
      author={Bo Liu and Yifeng Zhu and Chongkai Gao and Yihao Feng and Qiang Liu and Yuke Zhu and Peter Stone},
      year={2023},
      eprint={2306.03310},
      archivePrefix={arXiv},
      primaryClass={cs.AI},
      url={https://arxiv.org/abs/2306.03310}, 
}

@misc{intelligence2025pi05visionlanguageactionmodelopenworld,
      title={$\pi_{0.5}$: a Vision-Language-Action Model with Open-World Generalization}, 
      author={Physical Intelligence and Kevin Black and Noah Brown and James Darpinian and Karan Dhabalia and Danny Driess and Adnan Esmail and Michael Equi and Chelsea Finn and Niccolo Fusai and Manuel Y. Galliker and Dibya Ghosh and Lachy Groom and Karol Hausman and Brian Ichter and Szymon Jakubczak and Tim Jones and Liyiming Ke and Devin LeBlanc and Sergey Levine and Adrian Li-Bell and Mohith Mothukuri and Suraj Nair and Karl Pertsch and Allen Z. Ren and Lucy Xiaoyang Shi and Laura Smith and Jost Tobias Springenberg and Kyle Stachowicz and James Tanner and Quan Vuong and Homer Walke and Anna Walling and Haohuan Wang and Lili Yu and Ury Zhilinsky},
      year={2025},
      eprint={2504.16054},
      archivePrefix={arXiv},
      primaryClass={cs.LG},
      url={https://arxiv.org/abs/2504.16054}, 
}

@misc{liang2025draganddropllmszeroshotprompttoweights,
      title={Drag-and-Drop LLMs: Zero-Shot Prompt-to-Weights},
      author={Zhiyuan Liang and Dongwen Tang and Yuhao Zhou and Xuanlei Zhao and Mingjia Shi and Wangbo Zhao and Zekai Li and Peihao Wang and Konstantin Schürholt and Damian Borth and Michael M. Bronstein and Yang You and Zhangyang Wang and Kai Wang},
      year={2025},
      eprint={2506.16406},
      archivePrefix={arXiv},
      primaryClass={cs.LG},
      url={https://arxiv.org/abs/2506.16406},
}

@misc{zhou2025hypergoalnetgoalconditionedmanipulationpolicy,
      title={Hyper-GoalNet: Goal-Conditioned Manipulation Policy Learning with HyperNetworks}, 
      author={Pei Zhou and Wanting Yao and Qian Luo and Xunzhe Zhou and Yanchao Yang},
      year={2025},
      eprint={2512.00085},
      archivePrefix={arXiv},
      primaryClass={cs.RO},
      url={https://arxiv.org/abs/2512.00085}, 
}

@misc{ha2016hypernetworks,
      title={HyperNetworks}, 
      author={David Ha and Andrew Dai and Quoc V. Le},
      year={2016},
      eprint={1609.09106},
      archivePrefix={arXiv},
      primaryClass={cs.LG},
      url={https://arxiv.org/abs/1609.09106}, 
}

@inproceedings{beck2022hypernet,
  title={Hypernetworks in Meta-Reinforcement Learning},
  author={Jacob Beck and Matthew Jackson and Risto Vuorio and Shimon Whiteson},
  booktitle={Conference on Robot Learning},
  year={2022},
  url={https://api.semanticscholar.org/CorpusID:253018758}
}

@article{liang2025makeanagentgeneralizablepolicynetwork,
  title={Make-An-Agent: A Generalizable Policy Network Generator with Behavior-Prompted Diffusion},
  author={Yongyuan Liang and Tingqiang Xu and Kaizhe Hu and Guangqi Jiang and Furong Huang and Huazhe Xu},
  journal={ArXiv},
  year={2024},
  volume={abs/2407.10973},
  url={https://api.semanticscholar.org/CorpusID:271212603}
}

@article{rt22023arxiv,
  title={RT-2: Vision-Language-Action Models Transfer Web Knowledge to Robotic Control},
  author={Anthony Brohan and Noah Brown and Justice Carbajal and Yevgen Chebotar and Krzysztof Choromanski and Tianli Ding and Danny Driess and Kumar Avinava Dubey and Chelsea Finn and Peter R. Florence and Chuyuan Fu and Montse Gonzalez Arenas and Keerthana Gopalakrishnan and Kehang Han and Karol Hausman and Alexander Herzog and Jasmine Hsu and Brian Ichter and Alex Irpan and Nikhil J. Joshi and Ryan C. Julian and Dmitry Kalashnikov and Yuheng Kuang and Isabel Leal and Sergey Levine and Henryk Michalewski and Igor Mordatch and Karl Pertsch and Kanishka Rao and Krista Reymann and Michael S. Ryoo and Grecia Salazar and Pannag R. Sanketi and Pierre Sermanet and Jaspiar Singh and Anikait Singh and Radu Soricut and Huong Tran and Vincent Vanhoucke and Quan Ho Vuong and Ayzaan Wahid and Stefan Welker and Paul Wohlhart and Ted Xiao and Tianhe Yu and Brianna Zitkovich},
  journal={ArXiv},
  year={2023},
  volume={abs/2307.15818},
  url={https://api.semanticscholar.org/CorpusID:260293142}
}

@misc{reed2022GATO,
      title={A Generalist Agent}, 
      author={Scott Reed and Konrad Zolna and Emilio Parisotto and Sergio Gomez Colmenarejo and Alexander Novikov and Gabriel Barth-Maron and Mai Gimenez and Yury Sulsky and Jackie Kay and Jost Tobias Springenberg and Tom Eccles and Jake Bruce and Ali Razavi and Ashley Edwards and Nicolas Heess and Yutian Chen and Raia Hadsell and Oriol Vinyals and Mahyar Bordbar and Nando de Freitas},
      year={2022},
      eprint={2205.06175},
      archivePrefix={arXiv},
      primaryClass={cs.AI},
      url={https://arxiv.org/abs/2205.06175}, 
}

@article{kim24openvla,
    title={OpenVLA: An Open-Source Vision-Language-Action Model},
    author={{Moo Jin} Kim and Karl Pertsch and Siddharth Karamcheti and Ted Xiao and Ashwin Balakrishna and Suraj Nair and Rafael Rafailov and Ethan Foster and Grace Lam and Pannag Sanketi and Quan Vuong and Thomas Kollar and Benjamin Burchfiel and Russ Tedrake and Dorsa Sadigh and Sergey Levine and Percy Liang and Chelsea Finn},
    journal = {CoRL},
    year={2024}
}

@misc{Black20240AV,
      title={$\pi_0$: A Vision-Language-Action Flow Model for General Robot Control}, 
      author={Kevin Black and Noah Brown and Danny Driess and Adnan Esmail and Michael Equi and Chelsea Finn and Niccolo Fusai and Lachy Groom and Karol Hausman and Brian Ichter and Szymon Jakubczak and Tim Jones and Liyiming Ke and Sergey Levine and Adrian Li-Bell and Mohith Mothukuri and Suraj Nair and Karl Pertsch and Lucy Xiaoyang Shi and James Tanner and Quan Vuong and Anna Walling and Haohuan Wang and Ury Zhilinsky},
      year={2026},
      eprint={2410.24164},
      archivePrefix={arXiv},
      primaryClass={cs.LG},
      url={https://arxiv.org/abs/2410.24164}, 
}

@inproceedings{Bousmalis2023RoboCat,
  title={RoboCat: A Self-Improving Generalist Agent for Robotic Manipulation},
  author={Konstantinos Bousmalis and Giulia Vezzani and Dushyant Rao and Coline Devin and Alex X. Lee and Maria Bauz{\'a} and Todor Davchev and Yuxiang Zhou and Agrim Gupta and Akhil Raju and Antoine Laurens and Claudio Fantacci and Valentin Dalibard and Martina Zambelli and Murilo Fernandes Martins and Rugile Pevceviciute and Michiel Blokzijl and Misha Denil and Nathan Batchelor and Thomas Lampe and Emilio Parisotto and Konrad Zolna and Scott E. Reed and Sergio Gomez Colmenarejo and Jonathan Scholz and Abbas Abdolmaleki and Oliver Groth and Jean-Baptiste Regli and Oleg O. Sushkov and Tom Rothorl and Jos{\'e} Enrique Chen and Yusuf Aytar and David Barker and Joy Ortiz and Martin A. Riedmiller and Jost Tobias Springenberg and Raia Hadsell and Francesco Nori and Nicolas Manfred Otto Heess},
  year={2023},
  url={https://api.semanticscholar.org/CorpusID:259203978}
}

@article{jiang2023vima,
  title={VIMA: General Robot Manipulation with Multimodal Prompts},
  author={Yunfan Jiang and Agrim Gupta and Zichen Zhang and Guanzhi Wang and Yongqiang Dou and Yanjun Chen and Li Fei-Fei and Anima Anandkumar and Yuke Zhu and Linxi (Jim) Fan},
  journal={ArXiv},
  year={2022},
  volume={abs/2210.03094},
  url={https://api.semanticscholar.org/CorpusID:252735175}
}

@misc{zhang2024oneshot,
      title={One-Shot Imitation Learning with Invariance Matching for Robotic Manipulation}, 
      author={Xinyu Zhang and Abdeslam Boularias},
      year={2024},
      eprint={2405.13178},
      archivePrefix={arXiv},
      primaryClass={cs.RO},
      url={https://arxiv.org/abs/2405.13178}, 
}

@article{li2025metal,
  title={METAL: A Multi-Agent Framework for Chart Generation with Test-Time Scaling},
  author={Bingxuan Li and Yiwei Wang and Jiuxiang Gu and Kai-Wei Chang and Nanyun Peng},
  journal={ArXiv},
  year={2025},
  volume={abs/2502.17651},
  url={https://api.semanticscholar.org/CorpusID:276580569}
}

@article{guo2024prediction,
  title={Prediction with Action: Visual Policy Learning via Joint Denoising Process},
  author={Yanjiang Guo and Yucheng Hu and Jianke Zhang and Yen-Jen Wang and Xiaoyu Chen and Chaochao Lu and Jianyu Chen},
  journal={ArXiv},
  year={2024},
  volume={abs/2411.18179},
  url={https://api.semanticscholar.org/CorpusID:274306259}
}

@inproceedings{fastweight2016,
  title={Using Fast Weights to Attend to the Recent Past},
  author={Jimmy Ba and Geoffrey E. Hinton and Volodymyr Mnih and Joel Z. Leibo and Catalin Ionescu},
  booktitle={Neural Information Processing Systems},
  year={2016},
  url={https://api.semanticscholar.org/CorpusID:568305}
}

@misc{Verbancsics2013GenerativeNF,
      title={Generative NeuroEvolution for Deep Learning}, 
      author={Phillip Verbancsics and Josh Harguess},
      year={2013},
      eprint={1312.5355},
      archivePrefix={arXiv},
      primaryClass={cs.NE},
      url={https://arxiv.org/abs/1312.5355}, 
}

@article{Brock2017SMASHOM,
  title={SMASH: One-Shot Model Architecture Search through HyperNetworks},
  author={Andrew Brock and Theodore Lim and James M. Ritchie and Nick Weston},
  journal={International Conference on Learning Representations},
  year={2018},
  url={https://api.semanticscholar.org/CorpusID:3489117}
}

@article{knyazev2021parameter,
  title={Parameter Prediction for Unseen Deep Architectures},
  author={Boris Knyazev and Michal Drozdzal and Graham W. Taylor and Adriana Romero-Soriano},
  journal={ArXiv},
  year={2021},
  volume={abs/2110.13100},
  url={https://api.semanticscholar.org/CorpusID:239768239}
}

@article{Hyperrepresentations2022,
  title={Hyper-Representations as Generative Models: Sampling Unseen Neural Network Weights},
  author={Konstantin Sch{\"u}rholt and Boris Knyazev and Xavier Gir'o-i-Nieto and Damian Borth},
  journal={ArXiv},
  year={2022},
  volume={abs/2209.14733},
  url={https://api.semanticscholar.org/CorpusID:252595700}
}

@misc{Peebles2022,
      title={Learning to Learn with Generative Models of Neural Network Checkpoints}, 
      author={William Peebles and Ilija Radosavovic and Tim Brooks and Alexei A. Efros and Jitendra Malik},
      year={2022},
      eprint={2209.12892},
      archivePrefix={arXiv},
      primaryClass={cs.LG},
      url={https://arxiv.org/abs/2209.12892}, 
}

@article{soro2024diffusionbasedneuralnetworkweights,
  title={Diffusion-based Neural Network Weights Generation},
  author={Bedionita Soro and Bruno Andreis and Hayeon Lee and Song Chong and Frank Hutter and Sung Ju Hwang},
  journal={ArXiv},
  year={2024},
  volume={abs/2402.18153},
  url={https://api.semanticscholar.org/CorpusID:268041405}
}

@misc{wang2024neural,
      title={Neural Network Diffusion}, 
      author={Kai Wang and Dongwen Tang and Boya Zeng and Yida Yin and Zhaopan Xu and Yukun Zhou and Zelin Zang and Trevor Darrell and Zhuang Liu and Yang You},
      year={2024},
      eprint={2402.13144},
      archivePrefix={arXiv},
      primaryClass={cs.LG}
}

@article{jin2024conditional,
  title={Conditional LoRA Parameter Generation},
  author={Xiaolong Jin and Kai Wang and Dongwen Tang and Wangbo Zhao and Yukun Zhou and Junshu Tang and Yang You},
  journal={ArXiv},
  year={2024},
  volume={abs/2408.01415},
  url={https://api.semanticscholar.org/CorpusID:271693672}
}

@misc{wang2025rpg,
      title={Recurrent Diffusion for Large-Scale Parameter Generation}, 
      author={Kai Wang and Dongwen Tang and Wangbo Zhao and Konstantin Schürholt and Zhangyang Wang and Yang You},
      year={2025},
      eprint={2501.11587},
      archivePrefix={arXiv},
      primaryClass={cs.LG},
      url={https://arxiv.org/abs/2501.11587}, 
}

@inproceedings{hegde2025warpdworldmodelassisted,
  title={WARPD: World model Assisted Reactive Policy Diffusion},
  author={Shashank Hegde and Satyajeet Das and Gautam Salhotra and Gaurav S. Sukhatme},
  year={2024},
  url={https://api.semanticscholar.org/CorpusID:278960144}
}

@misc{zhou2025liberopro,
      title={LIBERO-PRO: Towards Robust and Fair Evaluation of Vision-Language-Action Models Beyond Memorization}, 
      author={Xueyang Zhou and Yangming Xu and Guiyao Tie and Yongchao Chen and Guowen Zhang and Duanfeng Chu and Pan Zhou and Lichao Sun},
      year={2026},
      eprint={2510.03827},
      archivePrefix={arXiv},
      primaryClass={cs.CV},
      url={https://arxiv.org/abs/2510.03827}, 
}

@misc{Li2025MetaVLAUM,
      title={MetaVLA: Unified Meta Co-training For Efficient Embodied Adaption}, 
      author={Chen Li and Zhantao Yang and Han Zhang and Fangyi Chen and Chenchen Zhu and Anudeepsekhar Bolimera and Marios Savvides},
      year={2026},
      eprint={2510.05580},
      archivePrefix={arXiv},
      primaryClass={cs.AI},
      url={https://arxiv.org/abs/2510.05580}, 
}

@misc{charakorn2025texttolora,
      title={Text-to-LoRA: Instant Transformer Adaption}, 
      author={Rujikorn Charakorn and Edoardo Cetin and Yujin Tang and Robert Tjarko Lange},
      year={2025},
      eprint={2506.06105},
      archivePrefix={arXiv},
      primaryClass={cs.LG},
      url={https://arxiv.org/abs/2506.06105}, 
}

@article{Achille2019Task2Vec,
  title={Task2Vec: Task Embedding for Meta-Learning},
  author={Alessandro Achille and Michael Lam and Rahul Tewari and Avinash Ravichandran and Subhransu Maji and Charless C. Fowlkes and Stefano Soatto and Pietro Perona},
  journal={2019 IEEE/CVF International Conference on Computer Vision (ICCV)},
  year={2019},
  pages={6429-6438},
  url={https://api.semanticscholar.org/CorpusID:60440365}
}

@article{james2018tecnets,
  title={Task-Embedded Control Networks for Few-Shot Imitation Learning},
  author={James, Stephen and Bloesch, Michael and Davison, Andrew J},
  journal={Conference on Robot Learning (CoRL)},
  year={2018}
}

@article{zhang2000flexible,
  title={A flexible new technique for camera calibration},
  author={Zhang, Zhengyou},
  journal={IEEE Transactions on pattern analysis and machine intelligence},
  volume={22},
  number={11},
  pages={1330--1334},
  year={2000},
  publisher={Ieee}
}

@InProceedings{HeLiu2021,
  author    = {Yanhao He and Steven Liu},
  booktitle = {2021 9th International Conference on Control, Mechatronics and Automation (ICCMA2021)},
  title     = {Analytical Inverse Kinematics for {F}ranka {E}mika {P}anda -- a Geometrical Solver for 7-{DOF} Manipulators with Unconventional Design},
  year      = {2021},
  month     = nov,
  publisher = {{IEEE}},
  doi       = {10.1109/ICCMA54375.2021.9646185},
}

@book{siciliano_textbook,
  title={Robotics: Modelling, Planning and Control},
  author={Siciliano, B. and Sciavicco, L. and Villani, L. and Oriolo, G.},
  isbn={9781846286414},
  lccn={2008939574},
  series={Advanced Textbooks in Control and Signal Processing},
  year={2010},
  publisher={Springer London}
}
